\documentclass[twoside]{article}

\usepackage[accepted]{aistats2026}  % use [anonymous] or [preprint] as needed

\usepackage{amsmath, amssymb, amsthm}
\usepackage{nicefrac} 
\usepackage{mathtools}
\usepackage{bm}
\usepackage{graphicx}
\usepackage{caption}
\usepackage{framed}
\usepackage{placeins}
\usepackage{subcaption}
\usepackage{algorithm}
\usepackage{balance}
\usepackage{algorithmic}
\usepackage{enumitem}
\usepackage{tikz}
\usepackage{color}
\usepackage{hyperref}
\usepackage{natbib}
\usepackage{booktabs}    
\theoremstyle{plain}
\newtheorem{theorem}{Theorem}[section]

\newtheorem{lemma}[theorem]{Lemma}

\theoremstyle{definition}

\newtheorem{assumption}[theorem]{Assumption}
\theoremstyle{remark}
\newtheorem{remark}[theorem]{Remark}
\DeclareMathOperator*{\argmin}{argmin}
\newcommand{\OCG}{\textsc{OCG}}
\newcommand{\CCV}{\textsc{CCV}}
\let\geq\geqslant
\let\leq\leqslant
\let\ge\geqslant

\mathcode`*="8000
{\catcode`*=\active
\gdef*{\star}}

\usepackage{xcolor}
\usepackage{hyperref}

\hypersetup{
    colorlinks=true,
    citecolor=blue,    % color for citations
    linkcolor=red,     % color for internal links
    urlcolor=cyan      % color for URLs
}

\allowdisplaybreaks
\setlist{leftmargin=*, noitemsep, topsep=2pt}

\newcounter{commentcounter}

\begin{document}

\twocolumn[

\aistatstitle{Projection-free Algorithms for Online Convex Optimization with Adversarial Constraints}

%\aistatsauthor{ Dhruv Sarkar~~~~~~ \And Aprameyo Chakrabartty~~~~~~~ \And Subhamon Supantha \And Palash Dey \And Abhishek Sinha }

%\aistatsaddress{ IIT Kharagpur \And IIT Kharagpur \And IIT Bombay \And IIT Kharagpur \And TIFR Mumbai, India }
\aistatsauthor{ Dhruv Sarkar \And Aprameyo Chakrabartty \And Subhamon Supantha }
\aistatsaddress{ IIT Kharagpur, India \And  IIT Kharagpur, India \And IIT Bombay, India }

\aistatsauthor{ Palash Dey \And Abhishek Sinha }
\aistatsaddress{ IIT Kharagpur, India \And TIFR Mumbai, India }
]
\runningauthor{Dhruv Sarkar, Aprameyo Chakrabartty, Subhamon Supantha, Palash Dey, Abhishek Sinha}
\mathcode`*="8000
{\catcode`*=\active
\gdef*{\star}}
\begin{abstract}
We study a generalization of the Online Convex Optimization (OCO) framework with time-varying adversarial constraints. In this setting, at each round, the learner selects an action from a convex decision set $\mathcal{X}$, after which both a convex cost function and a convex constraint function are revealed. The objective is to design a computationally efficient learning policy that simultaneously achieves low regret with respect to the cost functions and low cumulative constraint violation (\CCV) over a horizon of length $T$.
A major computational bottleneck in standard OCO algorithms is the projection operation onto the decision set $\mathcal{X}$. However, for many structured decision sets, linear optimization can be performed efficiently. Motivated by this, we propose a \emph{projection-free} online conditional gradient (OCG)-based algorithm that requires only a single call to a linear optimization oracle over $\mathcal{X}$ per round. Our approach improves upon the state of the art for projection-free online learning with adversarial constraints, achieving $\tilde{O}(T^{\nicefrac{3}{4}})$ bounds for both regret and CCV.

Our algorithm is conceptually simple. It constructs a surrogate cost function as a nonnegative linear combination of the cost and constraint functions, and feeds these surrogate costs into a novel adaptive online conditional gradient subroutine introduced in this paper. We further extend our framework to the bandit setting, where we show that a new form of surrogate loss is necessary to properly handle bandit feedback—an issue overlooked in prior work. Finally, we develop an efficient Follow-the-Perturbed-Leader (FTPL)-based algorithm, particularly well-suited for online combinatorial optimization problems with discrete actions, which also achieves $O(T^{\nicefrac{3}{4}})$ regret and CCV.
\end{abstract}
\section{Introduction} \label{intro}
Online Convex Optimization (OCO) provides a foundational framework for studying decision-making under uncertainty, especially in optimization and machine learning \citep{hazan2022introduction}. In this problem, an online algorithm competes against an adaptive adversary to minimize \emph{regret}, defined as the cumulative cost incurred by the algorithm and the cumulative cost incurred by the best-fixed action in hindsight, where the cost functions are assumed to be convex. 
%In this paper, we consider a generalization of this framework with additional adversarial constraints, which are revealed in an online fashion, along with the cost functions. 
We generalize this framework by incorporating adversarial constraints, where convex constraint functions are revealed online along with the cost functions.
This problem, commonly known as \emph{Constrained} Online Convex Optimization 
(COCO), has been extensively studied in the literature. Specifically,
on round $t\geq 1$, the online algorithm first selects a feasible action $x_t$ from a convex decision set $\mathcal{X}.$ After that, the adversary reveals a convex cost function $f_t :\mathcal{X} \mapsto \mathbb{R}$ and a convex constraint function $g_t: \mathcal{X} \mapsto \mathbb{R}.$ 
%As we will see in the sequel, our algorithm requires only first-order feedback, namely $\nabla f_t(x_t), \nabla g_t(x_t)$ and $g_t(x_t).$ 
As a result, the algorithm incurs an instantaneous cost of $f_t(x_t)$ and a constraint violation penalty (in case the current constraint is not satisfied) of $\max(0, g_t(x_t))\equiv g_t(x_t)^+.$ Let $T$ be the number of rounds for which the above game is played. The goal of the online policy is to choose a sequence of feasible actions $\{x_t\}_{t=1}^T$ so that both the regret ($\textsc{Regret}_T$) and the Cumulative Constraint Violations ($\textsc{CCV}_T$), are simultaneously minimized, where we define
\begin{eqnarray} \label{metrics}
	\textsc{Regret}_T(x^\star) &\stackrel{\textrm{(def.)}}{=}& \sum_{t=1}^T f_t(x_t)- \sum_{t=1}^Tf_t(x^\star), \nonumber\\
	\textsc{CCV}_T &\stackrel{\textrm{(def.)}}{=}& \sum_{t=1}^T g_t(x_t)^+. 
\end{eqnarray}
In the above definition, $x^\star \in \mathcal{X}$ is any fixed benchmark which satisfies all constraints, namely $g_t(x^\star) \leq 0, \forall t.$ We have also derived results while considering a stronger benchmark, one where it does not have to satisfy all the constraints and some violation is allowed. We have dealt with this in more detail in Appendix \ref{sec:budget_constraints}. The analysis and the definition of the benchmark follows from \citet{sarkar2025online}.

Note that when all constraint functions are identically equal to zero, the COCO problem reduces to the standard OCO problem, where the only objective is to minimize the regret. 
%The COCO problem has been extensively studied in the literature 
The COCO problem has been the subject of extensive study, including works by
\citet{sinha2024optimal, garber2024projectionfree, guo2022online, yi2021regret, mahdavi2012trading}. The optimal $O(\sqrt{T})$ regret and $\tilde{O}(\sqrt{T})$ CCV bound for the COCO problem was recently obtained by \citet{sinha2024optimal}, who proposed a first-order online algorithm using the AdaGrad subroutine. 
A major computational bottleneck of their algorithm, common to many standard OCO algorithms, \emph{e.g.,} Online Gradient Descent (OGD), is the Euclidean projection step onto the decision set
%A major computational bottleneck in many OCO algorithms, including Online Gradient Descent (OGD), is the Euclidean projection step onto the decision set
$\mathcal{X}$. This step requires solving a quadratic optimization problem over the decision set on every round and can be a computational bottleneck for high-dimensional problems, which are quite common in practice. 

To get around the above computational inefficiency, many efficient \emph{projection-free} algorithms have been proposed in the literature for the standard OCO problem, albeit with a worse regret guarantee. Instead of the projection step, these algorithms rely on a linear optimization step, which can be efficiently computed for many structured problems of interest, such as the shortest path problem on graphs, semidefinite optimization, optimization over matroid polytopes, optimal ranking problem, etc \citep{hazan2022introduction, garber2024projectionfree}.
\citet{hazan2012projection} proposed the first projection-free algorithm for the standard OCO problem, called Online Conditional Gradient (\textsc{OCG}), which achieves $O(T^{\nicefrac{3}{4}})$ regret. \citet{garber2022new} proposed a refined projection-free algorithm for the standard OCO problem, which achieves $O(T^{\nicefrac{3}{4}})$ regret for all intervals provided that a uniform upper bound to the Lipschitz constants of the cost functions is known \emph{a priori}. Furthermore, assuming access to an efficient membership oracle for the decision set, \citet{mhammedi2022efficient} proposed an OCO policy that achieves $O(\sqrt{T})$ regret with $\tilde{O}(T)$ many calls to the membership oracle. Coming back to the COCO problem, 
\citet{garber2024projectionfree} recently proposed a projection-free policy for the constrained OCO problem with $O(T^{\nicefrac{3}{4}})$ regret and $O(T^{\nicefrac{7}{8}})$ CCV. They proposed two projection-free algorithms, where one uses convex optimization while the other uses only linear optimization.
%The former achieves adaptive bounds that are valid for each time step, while the latter achieves bounds which are only valid for the entire sequence of costs and constraints. 
See Table \ref{comp-table} for a brief review of the known results on COCO and Section \ref{related} in the Appendix for more discussion on the related works. In this paper, we build upon the general regret decomposition framework developed by \citet{sinha2024optimal}. While their work establishes the foundational approach of using a surrogate loss to bound cumulative constraint violations, our transition to a projection-free setting introduces fundamental algorithmic and analytical challenges. Specifically, we propose a new Lipschitz-adaptive Online Conditional Gradient (OCG) policy. Because our step sizes dynamically adapt to the data-dependent Lipschitz constants of the surrogate losses, our regret bound inherently depends on a logarithmic ratio term involving $L_t/L_{t-1}$. To handle this price of adaptivity, which breaks the exponential potential function analysis used in prior projection-based work \citet{sinha2024optimal}, we introduce a novel power-law potential function and a specifically tailored initialization strategy. Through this new analysis, we achieve bounds of $\tilde{O}(T^{\nicefrac{3}{4}})$ for both regret and CCV in the full-information and bandit feedback settings.

%Further, we do so by only relying on linear optimization while retaining the adaptivity of our bounds.

We now highlight the major differences of the \OCG-based algorithm presented in this paper with the independent and concurrent work of \citet{wang2025revisitingprojectionfreeonlinelearning}, which addresses the same problem as ours.
While both studies achieve similar theoretical guarantees, they are methodologically distinct.
A key difference lies in how we address the unknown upper bound on the Lipschitz constant, $\tilde{G}$, for the surrogate loss functions.
\citet{wang2025revisitingprojectionfreeonlinelearning} employs the \textbf{doubling trick}, a general technique for handling unknown parameters in online learning.
This method partitions the time horizon into phases and restarts the entire algorithm with a revised estimate for $\tilde{G}$ for the next phase whenever the previous estimate is found to be small.
Such restarting techniques do not leverage the information accumulated in prior phases.
For a broader discussion on the pitfalls of the doubling trick, we refer the reader to \citet{zhang2024improving}, \citet{kwon2014continuous}, and \citet{luo2014towards}, where it has been argued that continuously adaptive algorithms perform better in practice.
In contrast, our work develops an \textbf{adaptive strategy} that continuously adjusts to the observed gradient norms \emph{without} requiring restarts.
This allows our algorithm to retain and build upon its accumulated knowledge throughout the entire time horizon.
%We believe this Lipschitz-adaptive approach is a valuable contribution to the online learning literature, offering a different perspective on handling unknown parameters.
The experiments in Section~\ref{expts} suggest that our Lipschitz-adaptive method offers distinctive practical performance gains over methods that rely on the doubling trick.

Furthermore, we identify critical technical issues in the bandit analysis presented by \citet{wang2025revisitingprojectionfreeonlinelearning}. Firstly, their analysis assumes $Q(t)$ (the CCV up to round $t$) in the bandit setting to be a deterministic quantity instead of a random variable, which is flawed as the actions in the bandit setting are randomized. More importantly, their surrogate loss function on round $t$ depends on the current action of the algorithm $x_t$ (see Eqn.\ (6) of \citep{wang2025revisitingprojectionfreeonlinelearning}). While this is permissible in the full-information setting, it is erroneous in the bandit setting, as without further assumptions, it would then be impossible to estimate the full gradient of the loss function through a randomized action. 
The latter is a subtle yet crucial point, and we provide a comprehensive discussion for the same in Section \ref{bandit_sec} and Section~\ref{wang_comparison} in the Appendix. Finally, we present a new $\mathsf{FTPL}$-based algorithm in 
Section \ref{ftpl} for linear cost and constraint functions.
\subsection{Our contributions} \label{contributions}
\begin{enumerate}
	\item We propose an online learning policy (Algorithm \ref{our_alg}) which makes exactly one call to an LP solver every round and yields $\tilde{O}(T^{3/4})$ regret and $\tilde{O}(T^{3/4})$ CCV bounds for COCO (Theorem \ref{perf_thm}). This result improves upon the bound of $O(T^{3/4})$ regret and $O(T^{7/8})$ CCV achieved by \citet{garber2024projectionfree} and matches that of the concurrent and independent work by \citet{wang2025revisitingprojectionfreeonlinelearning} without using the doubling trick.  
    %We also extend our policy to the bandit feedback setting with the same performance bounds.
	%In brief, our algorithm simply runs the adaptive \OCG~ policy on a surrogate cost function sequence.
	\item The proposed algorithm requires only first-order information of the cost and constraint functions (more precisely, $\nabla f_t(x_t), \nabla g_t(x_t),$ and $g_t(x_t)$ on round $t$ \footnote{For a (not necessarily differentiable) convex function $h,$ $\nabla h(x)$ refers to any arbitrary subgradient of $h$ at the point $x.$ All logarithms in this paper are taken w.r.t. the natural base.}), and hence, is efficient. We also extend our algorithm to the bandit feedback setting where only the values of the cost and constraint functions (\emph{i.e.,} $f_t(x_t)$ and $g_t(x_t)$) are revealed. Here we achieve state-of-the-art guarantees of $\tilde{O}(T^{\nicefrac{3}{4}})$ expected regret and  $\tilde{O}(T^{\nicefrac{3}{4}})$ expected violation. The bandit analysis also involves the emphasis on some novel and subtle points which have been overlooked in related works, and we provide a comprehensive discussion on the same for the sake of future works. Those ideas could be of independent interest.
    %Notably, we use the \textit{blocking} technique to divide the horizon into non-overlapping blocks which helps us control the variance of the gradient estimator and thus the regret and violation bounds. We also have to alter the analysis of \cite{sinha2024optimal} to manage some technical difficulties in the bandit setting.
	 \item  As a pre-requisite for our algorithm, we propose a Lipschitz-adaptive Online Conditional Gradient (OCG) algorithm for the standard OCO problem (Algorithm \ref{ocg_alg}), deriving the first known scale-free and Lipschitz adaptive regret bounds for OCG. This result might be of independent interest. Crucially, this result allows us to sidestep the use of the impractical doubling trick that related work relied on.
     %As a technical contribution, we propose an adaptive version of the Online Conditional Gradient (OCG) algorithm for the standard OCO problem and derive a scale-free adaptive regret bound for this algorithm. To the best of our knowledge, this is the first known adaptive regret bound for the OCG policy and is of independent interest. 
     %This also means that ours is the first projection-free COCO algorithm that gets adaptive bounds that hold for each round of the sequential process while relying only on linear optimization.
   %  \item We numerically simulate our proposed policy and two other benchmarks and compare their performance on a public dataset. 
   \item In the special case of online linear optimization, where both the cost and constraint functions are linear, we propose a Follow-the-Perturbed-Leader ($\mathsf{FTPL}$)-based policy that attains slightly improved regret and cumulative constraint violation (CCV) bounds of $O(T^{3/4})$ compared to the previous OCG-based algorithm. Moreover, for a broad class of combinatorial problems, such as shortest path and maximum spanning tree, $\mathsf{FTPL}$ offers significant computational advantages over $\mathsf{OCG}$. In particular, $\mathsf{FTPL}$ directly outputs a feasible combinatorial solution (e.g., a path or a spanning tree), whereas $\mathsf{OCG}$ requires the design of a non-trivial sampling procedure to extract such solutions from its (typically fractional) outputs.
 	\end{enumerate}

% \section{Related Work}
% In Table 1 we summarize the results of prior arts and also state their computational cost and the assumptions they make. In this section, we provide comparisons with two of our most closely related works - \cite{sinha2024optimal} and \cite{garber2024projectionfree}.

\section{Preliminaries: Adaptive OCO Algorithms} \label{ocg}

%\paragraph{Need for \emph{adaptive} projection-free OCO subroutines:} 
Our proposed online policy simply passes a sequence of surrogate cost functions (defined in Eqn.\ \eqref{surr-cost}) to an \emph{adaptive} version of the Online Conditional Gradient (\OCG) algorithm. The standard non-adaptive version of the \textsc{OCG} algorithm with a fixed step size requires a known uniform upper bound to the Lipschitz constants of all future cost functions \citep[Algorithm 26]{hazan2022introduction}. In this problem, we will see that the Lipschitz constants of the surrogate cost functions depend on the past actions of the algorithm. Thus, it is not possible to effectively upper bound the Lipschitz constants \emph{a priori}. 
%To get around this issue, in this section, we propose an extension of the standard \OCG ~algorithm, which makes use of adaptive step sizes and does not need any \emph{a priori} estimates of the Lipschitz constants. 
To address this issue, we propose an adaptive OCG algorithm that eliminates the need for a priori estimates of Lipschitz constants by leveraging adaptive step sizes.
In Section \ref{proj-free-coco}, we invoke this algorithm as a subroutine to construct an online policy for the Constrained OCO problem, which enjoys $\tilde{O}(T^{\nicefrac{3}{4}})$ regret and $\tilde{O}(T^{\nicefrac{3}{4}})$ Cumulative Constraint Violations (CCV) bounds. Our algorithm is conceptually simpler than \citet{garber2024projectionfree}, which makes use of an approximately feasible projection oracle. Furthermore, our algorithm achieves a better CCV bound (see Table \ref{comp-table}), which is also reflected in the numerical experiments (see Section \ref{expts}).
%Our proposed projection-free online algorithms critically make use of a class of OCO policies which do not need to know \emph{any} parameters of the future cost functions, including a uniform upper bound to their Lipschitz constants, yet achieve a similar regret bound when this knowledge is \emph{a priori} available.  

%Although the term \emph{adaptive} may be used in multiple contexts in the online learning literature, we use the term precisely in the above sense. 
While the term \emph{adaptive} is used in various contexts in online learning literature, here it specifically refers to the above definition.
A standard example of an adaptive OCO policy is \textsc{AdaGrad} \citep{duchi2011adaptive}, which was used by \citet{sinha2024optimal} in their COCO policy. Although \textsc{AdaGrad} enjoys a data-dependent adaptive regret bound \citep[Theorem 4.14]{orabona2019modern}, it is not projection-free. 
%We first propose an adaptive version of the online conditional gradient (\textsc{OCG}) algorithm. 
To the best of our knowledge, tight adaptive regret bounds for the \textsc{OCG} policy have not appeared in the literature before. Closest to this problem is the work by \citet[Theorem 9]{garber2022new}, who proposed a version of the OCG policy which achieves $O(T^{\nicefrac{3}{4}})$ regret over any time interval. However, their algorithm needs to know a uniform upper bound to the Lipschitz constants of the cost functions and, hence, is not adaptive in our sense.  As discussed previously, and also in section \ref{wang_comparison}, the algorithm proposed in \cite{wang2025revisitingprojectionfreeonlinelearning} does not make use of an adaptive OCG and instead runs multiple OCGs sequentially to account for the unknown Lipchitz constants, where each new OCG corresponds to a new estimate of the Lipchitz constant. This involves discarding the previous information whenever a new OCG run is started and thus is not very practically efficient (see Section \ref{expts}). 

%In the following, we revisit these algorithms and derive an adaptive regret bound for an adaptive version of \textsc{OCG}. 
%Unless stated otherwise, all norms in this paper will refer to the standard Euclidean $2$-norm and will be denoted by $||\cdot||.$ 
\subsection{Adaptive Online Conditional Gradient Algorithm} \label{ocg-section}
Recall that the ubiquitous online gradient descent (OGD) algorithm takes a gradient descent step on the last revealed cost function on every round and then projects the resulting (possibly infeasible) action back on the convex decision set $\mathcal{X}$ \citep[Algorithm 8]{hazan2022introduction}. OGD uses Euclidean projection, which could be expensive to compute on every round for many decision sets of practical interest (\emph{e.g.,} flow polytope on graphs). On the contrary, the Online Conditional Gradient Algorithm (\emph{a.k.a.} Online Frank-Wolfe) algorithm avoids the expensive projection step altogether and replaces it with a cheaper linear optimization step over the decision set \cite{hazan2012projection}. For many structured decision sets of practical interest, well-known and efficient subroutines exist that can perform linear optimization over the decision set efficiently. Standard examples include Dijkstra's algorithm for linear optimization over the flow polytope, the maximum weight matching algorithm for linear optimization over the Birkhoff-von Neumann polytope, and the greedy algorithm for linear optimization over the Matroid polytope.   \citet{hazan2012projection} showed that their proposed OCG algorithm achieves $O(T^{\nicefrac{3}{4}})$ regret for the OCO problem. In Algorithm \ref{ocg_alg}, we propose an adaptive version of the standard OCG policy.

%\cmt{Briefly explain why we need adaptive bounds in the first place. Also state the standard non-adaptive bound for the OCG algorithm.}
\begin{algorithm}
\caption{Adaptive Online Conditional Gradient} \label{ocg_alg}
\begin{algorithmic}[1]
    \STATE \textbf{Input} : Closed and convex decision set $\mathcal{X}$, Horizon length $T$,  sequence of convex cost functions $\{\hat{f}_t\}_{t=1}^{T},$ where the function $\hat{f}_t$ is $L_t$-Lipschitz, $\textrm{diam}(\mathcal{X}) = D$
    \STATE \textbf{Parameters}: $\eta_t = \frac{D}{2L_tT^{\frac{3}{4}}}, \sigma_t = \min \{1, \frac{2}{\sqrt{t}}\}, ~ t\geq 1 $ 
    %\State \textbf{Intitialization}: Set $x_1 \in \mathcal{X}, Q(0)=0$
    \STATE Initialize $x_1 \in \mathcal{X}$ arbitrarily 
    \FOR{$t=1:T$}
        \STATE Play $x_t,$ observe $\hat{f}_t,$ and compute $\nabla_t \equiv \nabla \hat{f}_t(x_t), L_t= || \nabla_t||,$ and $\eta_t$
        \STATE Define the regularized cumulative cost function
        \begin{eqnarray} \label{surr_fn}
        	   F_t(x)= \sum_{\tau=1}^{t} \nabla_\tau^{\top} x+\frac{\left\|x-x_1\right\|^2}{\eta_{t}}.
        \end{eqnarray} 
       % \STATE Using the LP oracle, efficiently compute 
       \STATE Solve a linear program over $\mathcal{X}$ to compute
        \begin{eqnarray} \label{v-def}
        v_t=\arg \min _{x \in \mathcal{X}}\left\{\nabla F_t\left(x_t\right) ^\top x\right\}.
             \end{eqnarray}
        \STATE Set \[x_{t+1} = (1-\sigma_t)x_t + \sigma_t v_t.\] 
    \ENDFOR

\end{algorithmic}
\end{algorithm}

The only difference between the standard non-adaptive version of the OCG policy, given in \citep[Algorithm 27]{hazan2022introduction}, and Algorithm \ref{ocg_alg} is that the fixed learning rate $\eta = \frac{D}{2LT^{\nicefrac{3}{4}}}$ of the former has been replaced with a sequence of adaptive learning rates $\eta_t= \frac{D}{2L_tT^{\nicefrac{3}{4}}}, ~t \geq 1,$ in the latter. Here $L_t$ is the Lipschitz constant of the $t$\textsuperscript{th} cost function $\hat{f}_t$. However, because of the adaptive regularization, this simple change requires non-trivial modifications in the statement and the proof of the regret bound, as given by the following theorem.
%
%\subsubsection{Regret Bound for the Adaptive Online Conditional Gradient Algorithm}
%%\cmt{Clearly define what $x^*_t$'s are. As of now the derivation does not make sense. }
%In this Section, we derive a data-dependent regret bound for an adaptive version of the standard OCG policy, where the learning rate $\eta_t$ varies inversely with the Lipschitz constant of the latest cost function $L_t$. 
%This result is of independent interest. 
\begin{theorem} \label{ocg-reg-thm}
Consider a sequence of convex cost functions $\{\hat{f}_t\}_{t \geq 1},$ where the function $\hat{f}_t$ is $L_t$-Lipschitz such that the Lipschitz constants $L_t$'s are monotone non-decreasing, \emph{i.e.,} $L_t\leq L_{t+1}, \forall t\geq 1$. 
    Then, the Online Conditional Gradient algorithm with  adaptive learning rates, described in Algorithm \ref{ocg_alg}, 
    %with the parameters $\eta_t= \frac{D}{2L_tT^{\frac{3}{4}}}, \sigma_t = \min \{1, \frac{2}{t^{\frac{1}{2}}} \}$, 
   achieves the following regret bound
    \begin{eqnarray} \label{ocg-reg-eq}
    &&\textrm{Regret}_T \equiv \sum\limits_{t=1}^{T} (\hat{f}_t(x_t) - \hat{f}_t(x^\star)) \nonumber\\
    &=&38 D L_T  T^{3/4} \sqrt{\log(Te L_T/L_1)})\bigg(1+\nonumber \\ && \frac{1}{304\sqrt{T}}(\max_{1\leq t \leq T} \frac{L_{t}}{L_{t-1}})\bigg).
    \end{eqnarray}  
    In particular, if $\max_{1\leq t \leq T} \frac{L_{t}}{L_{t-1}}$ is upper bounded by a constant, then the regret is dominated by the first term only. 
\end{theorem}
\begin{remark}
It is important to note that assuming the monotonicity of the Lipschitz constants incurs no loss of generality. One can always enforce this property by defining a new Lipschitz constant as $\tilde{L}_t:= \max_{1\leq \tau \leq t} L_\tau.$ In particular, due to the non-decreasing nature of the CCV variable $Q(t),$ the monotonicity condition naturally holds in the case of the surrogate cost function.
\end{remark}

\begin{remark}
  It is also worth emphasizing that the appearance of the ratio $\max_{1\leq t \leq T} \frac{L_{t}}{L_{t-1}}$ in the regret bound above results from the use of a non-proximal regularizer in the underlying FTRL subroutine. This is known as the so-called \emph{off-by-one} problem in the literature \citep{mcmahan2014analysis}.  
\end{remark}
\iffalse
\begin{eqnarray*}
	\sum_{t} \frac{L_t^2}{L_{t-1}} &=& \sum_{t} \frac{L_t^2-L_{t-1}^2}{L_{t-1}} + \sum_{t} L_{t-1} \\
	&\leq & \sum_t 2L_t \frac{L_t-L_{t-1}}{L_{t-1}} +TL_T\\
	&\leq & 2L_T  
\end{eqnarray*}
\fi

\iffalse
So putting it all together and substituting the value of $\eta_t$, 
\begin{equation}
    \text{Regret}_T\text{(OCG)} \leq \frac{4D}{T^{\frac{3}{4}}} \sum\limits_{t=1}^{T}L_t + 2\frac{\psi(x^*) - \psi(x_1)}{D}L_TT^{\frac{3}{4}} + \frac{D}{4L_TT^{\frac{3}{4}}}\sum\limits_{t=1}^{T}L_t^2
\end{equation}

Using $L_t \leq L_T$, $\psi(x^*)\leq D^2$ and $\psi(x_1) = 0$, we get

\begin{equation}\label{regdashbound}
    \text{Regret}_T\text{(OCG)} \leq 4DL_TT^{\frac{1}{4}} + 2DL_TT^{\frac{3}{4}}  + \frac{1}{4}DL_TT^{\frac{1}{4}} \leq 8DL_TT^{\frac{3}{4}}
\end{equation}
\fi
% \section{Regret of Online Conditional Gradient}

% According to Theorem 7.3 of Hazan(2022),  

% \begin{theorem}\label{ocgbound}
%     Online Conditional Gradient with parameters $\eta = \frac{D}{2GT^{\frac{3}{4}}}, \sigma_t=\min\{1,\frac{2}{t^{\frac{1}{2}}}\}$, attains the following guarantee
%     \[
%     \text{Regret}_T = \sum\limits_{t=1}^{T} f_t(x_t) - \min\limits_{x^* \in \mathcal{X}} \sum\limits_{t=1}^{T} f_t(x^*) \leq 8DGT^{\frac{3}{4}}
%     \]
%     where $D$ is the diameter of admissible set, $G$ is the Lipschitz constant of the functions $\{f_t\}$.
% \end{theorem}

\section{Projection-free COCO} \label{proj-free-coco}
%On round $t,$ the online algorithm first selects a feasible action $x_t$ from the decision set $\mathcal{X}.$ After that, the adversary reveals a cost function $f_t :\mathcal{X} \mapsto \mathbb{R}$ and a constraint function $g_t: \mathcal{X} \mapsto \mathbb{R}.$ As we will see in the sequel, our algorithm requires only first-order feedback, namely $\nabla f_t(x_t), \nabla g_t(x_t)$ and $g_t(x_t).$ As a result, the algorithm incurs an instantaneous cost of $f_t(x_t)$ and a constraint violation penalty of $\max(0, g_t(x_t))\equiv g_t(x_t)^+.$ The objective of the online policy is to choose a sequence of actions $\{x_t\}_{t\geq 1}$ so that both the \textsc{Regret} and the Cumulative Constraint Violations (\textsc{CCV}), defined below, are simultaneously minimized.  
%\begin{eqnarray} \label{metrics}
%	\textsc{Regret}_T(x^\star) &=& \sum_{t=1}^T f_t(x_t)-f_t(x^\star) \\
%	\textsc{CCV}_T &=& \sum_{t=1}^T g_t(x_t)^+. 
%\end{eqnarray}

%\cmt{Describe the problem briefly here.}
%\paragraph{Assumptions}
In this Section, we propose a projection-free online policy for the Constrained OCO (COCO) problem. 
We make the following assumptions common in the COCO literature \citep{neely2017online, guo2022online, sinha2024optimal, garber2024projectionfree}.
\begin{assumption}[Convexity and Lipschitzness]
The decision set $\mathcal{X}$ is closed and convex with a finite Euclidean diameter $D.$
	Furthermore, each of the cost and constraint functions are convex and $G$-Lipschitz, \emph{i.e.},$
		|f_t(x)-f_t(y)|\leq G ||x-y||,
		 |g_t(x)-g_t(y)| \leq G||x-y||,~\forall t, x,y \in \mathcal{X}.$
\end{assumption}

\begin{assumption}[Feasible Benchmark]
%	The regret with respect to the cost functions is computed against a fixed benchmark $x^\star \in \mathcal{X}$ which satisfies all constraints, 
Regret is measured relative to a fixed benchmark action $x^\star$ 
that satisfies all constraints
    \emph{i.e.,}  $g_t(x^\star) \leq 0, ~\forall t \in [T].$
\end{assumption}
The following computational assumption is specific to the projection-free literature, which puts an additional restriction on how the decision set $\mathcal{X}$ can be accessed by the algorithm \citep{garber2024projectionfree}. 
\begin{assumption}[LP Oracle for the decision set $\mathcal{X}$]
	The decision set $\mathcal{X}$ is accessible only through a linear optimization oracle $\mathcal{O},$ which takes a vector $v$ as input and returns $\arg \min_{x \in \mathcal{X}} \langle v, x \rangle.$%\footnote{It is implicitly assumed that the vectors $v$ and $x$ are of the same length.} 
	\end{assumption}
As discussed before, for many structured decision sets $\mathcal{X}$ of practical importance (\emph{e.g.,} flow-polytope), the oracle $\mathcal{O}$ can be implemented efficiently.

\subsection{Algorithm Design and Analysis} \label{analysis_sec}
Our proposed algorithm, described in Algorithm \ref{our_alg}, simply passes a sequence of surrogate cost functions to the adaptive OCG subroutine given in Algorithm \ref{ocg_alg}. 
%\footnote{To simplify the algebra, without any loss of generality, we assume that $T \geq e^2 \geq 8$.}
\begin{algorithm}
\caption{Projection-Free COCO}\label{our_alg}
\begin{algorithmic}[1]
\STATE \textbf{Inputs:} Convex decision set $\mathcal{X}$ of diameter $D,$ sequence of convex cost functions $\{f_t\}_{t \geq 1},$ and convex constraint functions $\{g_t\}_{t \geq 1}$ with Lipschitz constant $G$
\STATE \textbf{Parameters:} $\Phi(x)=x^p, V= (cp^2 T^{3/4})^p, p = \log T, c=144GD, Q(0)= GD \log T.$ 
\STATE Initialize $x_1 \in \mathcal{X}$ arbitrarily 
\FOR{$t=1:T$}
\STATE Play $x_t,$ observe $f_t(\cdot),$ and $g_t(\cdot),$
\STATE $Q(t)= Q(t-1)+ g_t(x_t)^+$
%\STATE Compute (the gradient of) the surrogate cost function $\hat{f}_t(\cdot)$ using equation \eqref{surr-cost}.  Pass $\hat{f}_t(\cdot)$ to Algorithm \ref{ocg_alg}.
\STATE Compute $\nabla \hat{f}_t(x_t)$ from Eqn.\ \eqref{surr-cost} and pass the (sub) gradient to Algorithm \ref{ocg_alg} 
\ENDFOR
\end{algorithmic}
\end{algorithm}

We now state our main result. The proof is given subsequently.
\begin{theorem} \label{perf_thm}
For $T \geq e^2 \geq 8$, Algorithm \ref{our_alg} achieves $\textrm{Regret}_T = \tilde{O}(T^{\nicefrac{3}{4}}), \textrm{CCV}_T = \tilde{O}(T^{\nicefrac{3}{4}}).$
\end{theorem}
\subsection{The Regret Decomposition Inequality} \label{gen_decomp}
Let $\Phi: \mathbb{R}_{\geq 0} \mapsto \mathbb{R}_{\geq 0}$ be a convex, differentiable, non-decreasing function, referred to as the \emph{Lyapunov} function. The function $\Phi(\cdot)$ will be used to control the CCV, which satisfies the following recursion:
\begin{eqnarray*}
	Q(t) = Q(t-1)+g_t(x_t)^+, ~~ t\geq 1. 
\end{eqnarray*}
%For technical reasons, which will become clear later, we initialize $Q(0) = GD \log T.$ 
For analytical convenience, $Q(0)$ is initialized as $GD \log T,$ a choice explained in subsequent sections. 
Thus $Q(T)$ constitutes an upper bound to $\textsc{CCV}_T,$ defined in \eqref{metrics}.
Using the convexity of the function $\Phi(\cdot),$ we have 
\begin{eqnarray*} 
	\Phi(Q(t-1)) \geq \Phi (Q(t)) + \Phi'(Q(t))(Q(t-1)-Q(t)).
\end{eqnarray*}
%Hence, we have the following upper bound for the change of the potential on round $t:$
Hence the increase of the Lyapunov function on round $t$ can be bounded as follows:
\begin{eqnarray} \label{drift_bd}
	\Phi(Q(t)) - \Phi(Q(t-1)) \leq \Phi'(Q(t)) g_t(x_t)^+.
\end{eqnarray}
We now fix a feasible action $x^\star \in \mathcal{X}$ so that $g_t(x^\star) \leq 0.$ Hence, from Eqn.\ \eqref{drift_bd}, we have 
\begin{eqnarray} \label{drift_bd2}
	&&\Phi(Q(t)) - \Phi(Q(t-1)) + Vf_t(x_t) -  V f_t(x^\star)  %\nonumber \\
	%&\leq& Vf_t(x_t) + \Phi'(Q(t)) g_t^+(x_t) -  V f_t(x^\star)\\
	\nonumber \\ \leq  &&Vf_t(x_t) + \Phi'(Q(t)) g_t^+(x_t) \nonumber \\ && -  V f_t(x^\star) - \Phi'(Q(t)) g_t^+(x^\star).
\end{eqnarray}
We now define the \emph{surrogate cost} function $\hat{f}_t$ for round $t$ as:
\begin{eqnarray} \label{surr-cost}
\hat{f}_t(x)= Vf_t(x)+ \Phi'(Q(t))g^+_t(x), ~~ x \in \mathcal{X},
\end{eqnarray}
where we have used the fact that since the adversary is allowed to be adaptive, the definition of the surrogate cost for round $t$ can depend on the action $x_t.$ 
%
%We now consider the following two cases.
%
%\paragraph{Case I: $g_t(x_t) \leq 0$:}
%We have \[\hat{f}_t(x_t) - \alpha \hat{f}_t(x^\star) = Vf_t(x_t) - \alpha V f_t(x^\star) = [\textrm{RHS}].\]
%\paragraph{Case II: $g_t(x_t) > 0$:}
%\[ \hat{f}_t(x_t) - \alpha \hat{f}_t(x^\star) = Vf_t(x_t) + \Phi'(Q(t)) g_t(x_t) - \alpha V f_t(x^\star) - \alpha \Phi'(Q(t)) g_t(x^\star) \geq [\textrm{RHS}],\]
%where in the last step we have used the feasibility of $x^\star$ and the fact that the Lyapunov function $\Phi(\cdot)$ is non-decreasing. 
Thus inequality \eqref{drift_bd2} takes the following form: 
\begin{eqnarray*}
		&&\Phi(Q(t)) - \Phi(Q(t-1)) + V\big(f_t(x_t) - f_t(x^\star)\big) \nonumber \\ &&\leq \hat{f}_t(x_t) - \hat{f}_t(x^\star).
\end{eqnarray*} 
Summing up the above, we arrive at the following Regret Decomposition Inequality which forms the basis for the subsequent analysis:
\begin{eqnarray} \label{gen-reg-decomp}
	\Phi(Q(t)) +V \textrm{Regret}_t \leq \Phi(Q(0))+\textrm{Regret}_t',
\end{eqnarray}
where $\textrm{Regret}_t$ on the LHS where $\textrm{Regret}'_t$ on the RHS correspond to the Regrets of the original and surrogate cost functions up to round $t$ respectively.
%\cmt{Write the algorithm in an algorithm environment. It should be just two steps - define the surrogate costs and then invoke the OCG policy. Refer to this policy throughout the paper (which we now refer to as "our algorithm").}

% of the policy up to round $t,$ \emph{i.e,,}
%\begin{eqnarray*}
%	\textrm{Regert}_T(\alpha) \equiv \sum_{t=1}^T \big(f_t(x_t) - \alpha f_t(x^\star)\big).
%\end{eqnarray*}
%Similarly, the RHS of \eqref{gen-reg-decomp} refers to the $\alpha$-Regret of the policy for the surrogate cost function sequence defined in Eqn.\ \eqref{surr-cost}. Note that in deriving the generalized regret decomposition inequality \eqref{gen-reg-decomp}, we did not place \emph{any} restriction on the sequence of cost or constraint functions. Hence, \eqref{gen-reg-decomp} is valid for even non-convex functions as well. 
%

\subsection{Regret and CCV Bound}
\label{regret_and_ccv}
%\cmt{$t$ and $T$'s are all messed up!}
%Recall that the $t$\textsuperscript{th} surrogate cost function is defined as 
%\[\hat{f}_t(x)= Vf_t(x)+ \Phi'(Q(t))g^+_t(x).\]
Since the cost and constraint functions are assumed to be $G$-Lipschitz, from Eqn.\ \eqref{surr-cost}, it follows that the surrogate cost $\hat{f}_t$ is an $L_t$-Lipschitz convex function where 
\begin{eqnarray} \label{lipschitz-const}
L_t = G(V+\Phi'(Q(t)).	
\end{eqnarray}
Furthermore, since $\Phi(\cdot)$ is convex and $Q(t)$ is monotone non-decreasing, it follows that $\{L_t\}_{t\geq 1}$ is also a monotone non-decreasing sequence.  
%Note that the surrogate loss functions are all $L_T$-Lipschitz continuous. Where, $L_T = (2D)^{-1}(V + \Phi'(Q(T)))$.
From Eqn.\ \eqref{gen-reg-decomp}, we have
\begin{equation} \label{reg-decomp}
    \Phi(Q(T)) + V\text{Regret}_T  \leq \Phi(Q(0))+ \text{Regret}'_T
\end{equation}
We choose a power-law potential function of the form $\Phi(x)=x^p$ with $p = \log T$. From Lemma \ref{ratio-bd}, we have $\max_{1\leq t \leq T} \frac{L_{t}}{L_{t-1}} \leq 1+e.$ Hence, the regret of the OCG policy used on the surrogate cost functions, given by Theorem \ref{ocg-reg-thm}, can be upper bounded as: 
%\cmt{constant needs to be changed}
\[ \textrm{Regret}_T' \leq  48 D L_T  T^{3/4} \sqrt{\log(Te L_T/L_1)}).\]
Substituting for the expression of $L_T$ from Eqn.\ \eqref{lipschitz-const} and 
using Lemma \ref{sqrt-ratio}, stated below, we can further simplify the above regret bound as: 
\[ \textrm{Regret}_T' \leq  144 GD T^{3/4} (V+pQ^{p-1}(T)) \log T.\]
%where we have used the expression for $L_T$ given by Eqn.\ \eqref{lipschitz-const}. 
%and used the fact that $\sqrt{x+y+z} \leq \sqrt{x}+\sqrt{y}+\sqrt{z}$. 
%\subsection{Taking the polynomial potential function}
Combining the above regret bound with Eqn.\ \eqref{reg-decomp} and letting $c \equiv 144GD,$ we arrive at the following inequality which will be used for simultaneously bounding the Regret and CCV: 
\begin{equation} 
\label{main-eq}
        Q^p(T)+V\textrm{Regret}_T\leq (GDp)^p+ cpT^{\frac{3}{4}}(V+pQ^{p-1}(T)),
\end{equation}
where we have used the fact that $p=\log T.$
\paragraph{Analysis:}
From the Lipschitzness of the cost functions, we trivially have $\textrm{Regret}_T \geq -GDT.$ This implies that 
\begin{align*}
 Q^p(t)&\leq 2cVT+(GDp)^p+cp^2T^{\frac{3}{4}}\,Q^{p-1}(T) \\
&\leq 2\max(2cVT+(GDp)^p,cp^2 T^{\frac{3}{4}}\,Q^{p-1}(T)).
\end{align*}
Hence,
\begin{equation}
\label{violation-bound}
 Q(T)\leq \max((4cVT)^{\frac{1}{p}} + 2GDp,2cp^2T^{\frac{3}{4}}).
\end{equation}
The above inequality bounds the CCV. To bound the Regret, 
using Eqn. \eqref{main-eq} once again, we have  
\begin{align} \label{reg-bd-expr}
&\textrm{Regret}_T \nonumber \\
&\leq cpT^{\frac{3}{4}} + \frac{(GDp)^p}{V}+ \frac{cp^2T^{\frac{3}{4}}Q^{p-1}(T)-Q^p (T)}{V} \nonumber  \\
&\stackrel{(b)}{\leq} cT^{\frac{3}{4}}\log\,T+\frac{(GDp)^p}{V} + \frac{1}{V}(cp^2T^{\frac{3}{4}})^p.
\end{align}
% and (b) by optimising the 2nd term of the RHS of the expression preceding it.
 Inequality (b) follows by considering the possible cases $Q(T) > cp^2T^{\nicefrac{3}{4}}$ and $0\leq Q(T) \leq cp^2T^{\nicefrac{3}{4}}$ and upper bounding the last term separately in each case. 
 Finally, we set $V=(cp^2T^{\frac{3}{4}})^p$ and recall that $p=\log\,T.$ Substituting the above into  \eqref{violation-bound} and \eqref{reg-bd-expr}, and simplifying the algebraic expressions, we obtain $\tilde{O}(T^{\nicefrac{3}{4}})$ bounds for both Regret and CCV, which proves Theorem \ref{perf_thm}.

\iffalse
\begin{equation}
    Q^m(T)+V\textrm{Regret}_T\leq cT^{\frac{3}{4}}(V+mQ^{m-1}(T))m\,\log\,T 
\end{equation}
\[
Q^m(t)\leq2VT+m^2T^{\frac{3}{4}}\log\,T\,Q^{m-1}(T)
\]
\[
\leq 2\max(2VT,m^2 T^{\frac{3}{4}}\log\,T\,Q^{m-1}(T))
\]
\[
Q(T)\leq O(\max((VT)^{\frac{1}{m}},m^2T^{\frac{3}{4}}\log\,t)
\]
Again from equation ,
\[
R_T\leq mT^{\frac{3}{4}}\log\,T +\frac{m^2T^{\frac{3}{4}}\log T\,Q^{m-1}(T)-Q^m (T)}{V}
\]
\[
\leq O(mT^{\frac{3}{4}}\log\,T+\frac{1}{V}(m^2T^{\frac{3}{4}}\log\,T)^m)
\]
Set $V=(m^2T^{\frac{3}{4}}\log\,T)^m$,\,$m=\log\,T.$ This yields $(VT)^{\frac{1}{m}}=(\log\,T)^3\,T^{\frac{3}{4}}$.
\fi
\iffalse
To complete the analysis, we state the following two supporting Lemmas, which were used in the above argument.
\begin{lemma} \label{ratio-bd}
Choosing $\Phi(x)=x^m$ with $m =\log T$ and for any $V>0,$ we have 
	\begin{eqnarray*}
		\max_{1\leq t \leq T} \frac{L_{t}}{L_{t-1}} \leq 1+e.
	\end{eqnarray*}
\end{lemma}
	See Section \ref{ratio-bd-proof} in the Appendix for the proof of this Lemma.

	\begin{lemma}
    \label{sqrt-ratio}
   Choosing $\Phi(x)=x^m$ with $m =\log T$ and for any $V>0,$ we have 
	\begin{eqnarray*}
		\sqrt {\log \frac{L_{T}}{L_{1}}} \leq \log T.
	\end{eqnarray*}
\end{lemma}
See Section \ref{sqrt-ratio-proof} in the Appendix for the proof.
\fi
%\input{algo}
\section{Projection-Free CBCO} \label{bandit_sec}

\newcommand{\ball}{\mathcal{B}}
\newcommand{\reals}{\mathbb{R}}
In this section, we extend our projection-free algorithm to the bandit-feedback setting against an oblivious adversary. This setting has previously been explored in both \citet{garber2024projectionfree} and \citet{wang2025revisitingprojectionfreeonlinelearning} where they proposed projection-free algorithms for constrained bandit convex optimization (CBCO). \cite{garber2019improvedregretboundsprojectionfree} achieved an expected regret bound of $\tilde{O}(T^{\nicefrac{3}{4}})$ and an expected CCV bound of $\tilde{O}(T^{\nicefrac{7}{8}})$. The results of \citet{wang2025revisitingprojectionfreeonlinelearning} do not actually hold and we have discussed the same in Appendix \ref{wang_comparison}. We achieve expected regret bound of 
$\tilde{O}(T^{\nicefrac{3}{4}})$ and expected CCV bound of $\tilde{O}(T^{\nicefrac{3}{4}})$.

As in \cite{flaxman2004online} we make the standard assumption that the feasible set $\mathcal{X}$ is full dimensional, contains the origin, and that there exists scalars $r,R>0$ such that $r\ball^n\subseteq\mathcal{X}\subseteq{}R\ball^n$, where $\ball^n$ denotes the unit Euclidean ball centered at the origin in $\reals^n$. Further note that $D=2R.$ Also, as in most bandit convex optimization literature, we assume that we have an oblivious adversary, that is, the cost and constraint functions are chosen ahead of the beginning of the play and are not affected by the randomness of the play.

The key algorithmic distinction in the bandit setting is that, at each round, the decision-maker observes only the incurred cost and constraint violation for the chosen action, rather than the entire cost and constraint functions. Consequently, one cannot compute exact gradients and must instead rely on gradient estimators whose bias and variance affect overall performance. To control the variance we use the \textit{blocking} technique and to ensure that unbiased estimation of gradient is possible, we change the form of the surrogate loss.  More details regarding the choice of estimator and notions like smoothing are given in Section \ref{bandit-background}.

\subsection{Adaptive Online Conditional Gradient under Bandit Feedback}

\begin{algorithm}
\caption{Adaptive Block Bandit Conditional Gradient}\label{alg:VRBCO}
\begin{algorithmic}[1]
    \STATE \textbf{Input} : Closed and convex decision set $\mathcal{X}$, Horizon length $T$,  sequence of convex cost functions $\{\hat{f}_t\}_{t=1}^{T},$ where the function $\hat{f}_t$ is $L_t$-Lipschitz and has sup-norm of $M_t$, $\textrm{diam}(\mathcal{X}) = D$ 
    \STATE \textbf{Parameters}: $\eta_m = \frac{D}{(\sqrt{n}M_{mK}+L_{mK})T^{\frac{3}{4}}}, ~ m\geq 1 $, $\epsilon = 16R^2T^{-1/2}\sqrt{\log T}$ 
    %\State \textbf{Intitialization}: Set $x_1 \in \mathcal{X}, Q(0)=0$
    \STATE Initialize $x_1 \in \mathcal{X}$ arbitrarily 
    \FOR{$m=1:\frac{T}{K}$}
        \FOR{$s = 1,.....,K$}
        \STATE $t = (m-1)K + s$ and $u_t \sim \mathcal{S}^n$
        \STATE Play $y_t \xleftarrow{} x_m + \delta u_t$, observe $\hat{f}_t(y_t)$ and compute $\nabla_{m,s} \equiv \frac{n\hat{f}_t(y_t)}{\delta}$
        \ENDFOR
        \STATE $\nabla_{m} \equiv \sum_{s=1}^{K} \nabla_{m,s}$
        \STATE Define the regularized cumulative cost function
        \begin{eqnarray} \label{surr_fn_bandit}
        	   F_m(x)= \sum_{\tau=1}^{m} \nabla_\tau^{\top} x+\frac{\left\|x-x_1\right\|^2}{\eta_{m}}.
        \end{eqnarray} 
        \STATE Run Algorithm \ref{alg:Cg} with set $\mathcal{X}_{\delta}$,  tolerance $\epsilon$, initial vector $x_{m}$, and function $F_{m}(x)$. 
        \STATE $x_{m+1} \xleftarrow{} \text{Output of Algorithm 3}$
    \ENDFOR

\end{algorithmic}
\end{algorithm}

Like in the full information setting, we would have to design an adaptive algorithm for the unconstrained setting to which the surrogate loss function would be passed. Therefore, the first step is to design something akin to an adaptive OCG for the bandit setting.

To keep the estimator’s variance under control, a naive extension of the OCG algorithm is insufficient. Instead, we employ a \emph{blocking} technique: the time horizon is divided into disjoint blocks of length~\(K\), and the algorithm holds a single action fixed throughout each block~\(m\), updating only at the beginning of block~\(m+1\). By choosing~\(K\) appropriately, this approach balances variance reduction against any block-wise regret accumulation. We call the resulting method \emph{blocked-OCG}, and we present it as Algorithm~\ref{alg:VRBCO}. We assume without loss of generality that $T$ is perfectly divisible by $K$.

It is important to note that the blocking technique has been used before \citep{hazan2020faster,garber2019improvedregretboundsprojectionfree, garber2024projectionfree, 1013135}. However, we mention it here to highlight the distinction between our approach to handling CBCO as opposed to the COCO at an algorithmic level. 

%Note that while in our adaptive algorithm in the full information setting the adaptivity manifested itself in form of the time-varying Lipchitz constants $L_t$, in this case we also care about the time-varying sup-norms $M_t$ of the loss functions. The introduction of the blocking technique introduced some more technical challenge in deriving the expected regret bound for this adaptive algorithm. 

The above algorithm calls the subroutine Algorithm \ref{alg:Cg} as a subroutine at the end of every block and the goal of the subroutine is to approximately optimize $F_m$ using LOO calls. The error tolerance $\epsilon$ is a measure of our proximity to the exact solution. The subroutine works by terminating when the \textit{Frank-Wolfe gap} is less than $\epsilon.$ Till the time that a low enough Franke-Wolfe gap is achieved, we keep making LOO calls and improving our decision. More details about Algorithm \ref{alg:Cg} is given in section \ref{cond-grad-sec}.

Note that we are making multiple calls at the end of each block. Thus, we also have to bound the expected total number of LOO calls at the end of $T$ rounds. 

We present Theorem \ref{thm:main-bandit} which presents the bound on both the expected regret and the expected number of LOO calls of the above algorithm.

Note that $N_m$ is the number of LOO calls made at the end of block $m.$ Further we define $G_t = \sqrt{n}M_t + L_t$ for the sake of notational convenience. Where $M_t$ and $L_t$ are the sup-norm and Lipchitz constant of the $t$-th loss respectively.

\begin{theorem}  \label{thm:main-bandit}
    %Let $L_m$ the number of calls in block $m$ to the linear optimization oracle. 
    Setting $\eta_m = \frac{D}{G_{mK}} T^{-\frac{3}{4}}$, $\delta = \sqrt{n}T^{-\frac{1}{4}} $, $K = \sqrt{T}$ in Algorithm \ref{alg:VRBCO}, guarantees that the expected regret is upper-bounded by
    {\small\begin{align}
        \mathbb{E}[\textrm{Regret}_{T}] \leq & 4\sqrt{n}L_TT^{\frac{3}{4}}+3L_TT\sqrt{\epsilon} \nonumber \\ &+ DG_T\big(1 + \max_{1\leq m\leq T/K} \frac{G_{mK}}{G_{(m-1)K}}\big) T^{3/4} , \nonumber
    \end{align}}
    and that the expected overall number of calls to the linear optimization oracle is upper-bounded by
    \begin{align}
        \mathbb{E} \left[ \sum_{m=1}^{\frac{T}{K}} N_m \right] \leq & ~ \frac{16R^4}{\epsilon^2} \left ( \frac{\sqrt{\epsilon}}{R} T^{1/4} + 1 + \sqrt{\log \frac{G_T}{G_1}}\right) . \nonumber
    \end{align}
\end{theorem}

The proof has been deferred to Appendix \ref{thm-proof}.

To achieve an amortized budget of one Linear Optimization Oracle (LOO) call per round (for a total of $T$ calls), we select an appropriate value for the error tolerance $\epsilon$. This choice is crucial as it governs the trade-off between the total number of oracle calls and the final regret bound.

\subsection{The Regret Decomposition Inequality}
One may be tempted to extend the approach in the full information setting and construct the surrogate loss as in there and pass that surrogate loss to an analogous adaptive bandit algorithm. This is essentially what \cite{wang2025revisitingprojectionfreeonlinelearning} tried to do. However, that would not be a correct approach. This is because doing so would break one of the most fundamental assumptions in bandit convex optimization - that we would be able to compute an unbiased estimate of the loss function. 

Consider that we construct a surrogate loss of the form $\hat{f}_t(x_t) = Vf_t(x_t) + \Phi'(Q(t))g_t^+(x_t).$ Notice that, the definition of the function $\hat{f}_t$ involves $Q(t)$ in it which in itself contains $x_t$. For there to be an unbiased estimate of $\hat{f}_t$, we need to at least make sure that the loss at time $t$ does not depend on the action at time $t$. Ultimately, this is the reason that it is assumed that we have an oblivious adversary as opposed to an adaptive one in the bandit setting. An adaptive adversary in the bandit setting would have the power to decide the loss function based on the action of the player and thus unbiased estimation of gradient would be impossible. In contrast, for an oblivious adversary, the sequence of loss functions does not depend on the randomness of play. However, unbiased estimation is possible even if the loss function depends on the randomness of play as long as it does not depend on the latest action $x_t.$ Thus, the loss function can depend on all the actions $x_{<t}$ but not on $x_t$ itself. Accordingly we modify the surrogate loss function, to be of the following form --
\begin{equation}
\label{bandit-surr-cost}
    \hat{f}_t(x_t) = Vf_t(x_t) + \Phi'(Q(t-1)+GD)g_t^+(x_t)
\end{equation}
In summary, we can say that the above equation implies that the surrogate loss $\hat{f}_t$ is $\mathcal{F}_{t-1}$ measurable, where $\{\mathcal{F}_{\tau}\}_{\tau \geq 1}$ is the natural filtration.

To derive the regret decomposition inequality, we would utilise the fact that the quantity $Q(t-1)+GD$ is an overestimate of $Q(t)$ because $g_t^+(x_t) \leq GD$.

\begin{algorithm}
\caption{Projection-Free CBCO}\label{our_alg_bandit}
\begin{algorithmic}[1]
\STATE \textbf{Inputs:} Convex decision set $\mathcal{X},$ sequence of convex cost functions $\{f_t\}_{t \geq 1},$ and convex constraint functions $\{g_t\}_{t \geq 1}.$
\STATE \textbf{Parameters:} $Q(0) = GD\sqrt{T}\log T, \Phi(x) = x^{\log T}$, $V = \big((30 + \frac{R}{r})\sqrt{n}(\log T)^{1/4}T^{3/4}\big)^{\log T}$
\STATE Initialize $x_1 \in \mathcal{X}$ arbitrarily 
\FOR{$t=1:T$}
\STATE Play $x_t,$ observe $f_t(x_t),$ and $g_t(x_t),$
\STATE Compute $\hat{f}_t(x_t)$ using equation \eqref{bandit-surr-cost}.  Pass $\hat{f}_t(x_t).$ to Algorithm \ref{alg:VRBCO}.
\ENDFOR
\end{algorithmic}
\end{algorithm}
\begin{align*} 
	\Phi(Q(t)) - \Phi(Q(t-1))  &\leq  \Phi'(Q(t)) g_t(x_t)^+ \nonumber \\ &\leq \Phi'(Q(t-1)+GD) g_t(x_t)^+
\end{align*}
\begin{align*} 
&\Phi(Q(t)) - \Phi(Q(t-1)) + Vf_t(x_t) -  V f_t(x^\star)  %\nonumber \\
	%&\leq& Vf_t(x_t) + \Phi'(Q(t)) g_t^+(x_t) -  V f_t(x^\star)\\
	\nonumber \\ \leq  &Vf_t(x_t) + \Phi'(Q(t-1)+GD) g_t^+(x_t) \nonumber \\ & -  V f_t(x^\star) - \Phi'(Q(t-1)+GD) g_t^+(x^\star).
\end{align*}
Summing up inequalities for $1\leq t \leq T$ and telescoping, we conclude that
\begin{align*}
\Phi(Q(T)) - \Phi(Q(0)) + V\textrm{Regret}_T \leq \textrm{Regret}_T'
\end{align*}
Taking expectation on both sides, 
\begin{align} 
\label{bandit-reg-decomp}
\mathbb{E}\Phi(Q(T)) - \Phi(Q(0)) + V\mathbb{E}\textrm{Regret}_T 
\leq\mathbb{E}\textrm{Regret}_T' 
\end{align}

Now we have got something analogous to the regret decomposition inequality for the full-information setting. The only difference is that we are appropriately taking expectation of the regret and violation over the randomness of the play. 

$\mathbb{E}\textrm{Regret}_T'$ is the expected regret incurred by the bandit algorithm when we pass the surrogate loss in eqn. \eqref{bandit-surr-cost} to it. We have shown that in Algorithm \ref{our_alg_bandit}.

Note that although the surrogate loss \eqref{bandit-surr-cost} on round $t$ depends on the actions of the online algorithm up to round $t-1,$ the benchmark $x^\star$, as discussed above, is oblivious to the action of the algorithms and can be decided at the start of the play because $\{f_t\}_{t\geq 1}$ and $\{g_t\}_{t\geq 1}$ are known apriori by our oblivious adversary. Indeed, the benchmark $x^\star$ would be chosen such that the cumulative loss of $\{f_t\}_{t\geq 1}$ is minimized while ensuring feasibility on $\{g_t\}_{t\geq 1}$.

Since the action of the algorithm at round $t$ is independent of the losses at round $t,$ it is clear that we can upper bound the regret of the surrogate problem against the benchmark $x^\star$ using any oblivious regret bound. Therefore, we take the help of Theorem \ref{thm:main-bandit} to help us bound $\mathbb{E}\textrm{Regret}_T'$ by using the form of the surrogate cost. 
\newcommand{\E}{\mathbb{E}}

\subsection{Expected Regret and Violation}
\begin{theorem} \label{perf_thm_bandit}
Algorithm \ref{our_alg_bandit} achieves $\E\textrm{Regret}_T = \tilde{O}(T^{\nicefrac{3}{4}}), \E\textrm{CCV}_T = \tilde{O}(T^{\nicefrac{3}{4}})$ while using an expected total number of less than $T$ LOO calls.
\end{theorem}

The proof has been deferred to Appendix \ref{bandit-reg-vio}.

The key feature of the proof is that using the form of the surrogate loss, we can bound $\sqrt{\log \frac{G_T}{G_1}}$ and then set $\epsilon = 16R^2T^{-1/2}\sqrt{\log T}$ to make the expected number of LOO calls less than $T$. From that, we get the bound on $\mathbb{E}\textrm{Regret}_T'$ as,
\begin{align*}
    \E\textrm{Regret}'_T \leq  c(V+ \E\Phi'(Q(T)))T^{3/4}
\end{align*}
where $c = (30 + \frac{R}{r}) \sqrt{n} (\log T)^{1/4}$ and we then plug this in the regret decomposition inequality \eqref{bandit-reg-decomp}, alongside some manipulations with the Jensen's inequality to get our desired bounds.

%\begin{align*}
%    &\mathbb{E}\Phi(Q(T)) - \Phi(Q(0)) + V\mathbb{E}\textrm{Regret}_T \\ &\leq c(V+ \E\Phi'(Q(T)))T^{3/4} 
%\end{align*}
 The takeaway from our bandit analysis should be that while the extension to CBCO from COCO is quite simple from an algorithmic point of view, the subtleties associated with the form of the surrogate loss that we assume makes the analysis non-trivial. As demonstrated in \cite{wang2025revisitingprojectionfreeonlinelearning} it is quite easy to miss this point and this further highlights our contribution. The points raised are of independent interest even beyond the projection-free setting. In particular, they will be important whenever one tries to use a reduction to BCO to analyse a CBCO problem along the lines of \cite{sinha2024optimal}.

\section{An Efficient FTPL-based Algorithm in the Full-Information Setting} \label{ftpl}

There exist efficient linear programming (LP) subroutines for many combinatorial optimization problems, such as shortest path, maximum matching, and maximum-weight spanning tree. While the $\mathsf{OCG}$ algorithm described in the previous sections can, in principle, be applied to the online versions of these problems, by convexifying the feasible set and designing an appropriate sampling procedure, this approach is indirect and often computationally inefficient.

In contrast, the Follow-the-Perturbed-Leader ($\mathsf{FTPL}$) algorithm requires solving only a single LP at each round and directly outputs a feasible solution (e.g., a path, a matching, or a spanning tree), thereby completely bypassing the need for sampling \citep{abernethy2016perturbation}. In this section, we show that $\mathsf{FTPL}$ can be used to solve constrained online linear optimization problems while achieving the same regret and cumulative constraint violation (CCV) guarantees as $\mathsf{OCG}$.

\paragraph{Setup.}

We consider a constrained online linear optimization (OLO) problem over an arbitrary compact (not necessarily convex) set $\mathcal{X}$. For example, $\mathcal{X}$ could be given by the set of all paths, or matchings, or spanning trees of some graph. At each round $t\geq 1$, the learner observes a linear cost function $f_t(x) = \langle f_t, x \rangle$ and a linear constraint function $g_t(x) = \langle g_t, x \rangle$, where $\|f_t\|, \|g_t\| \leq 1$ for all $t \geq 1$. For simplicity, we assume that both sequences are generated by an oblivious adversary. Unlike the previous section, we allow the learning algorithm to be randomized. The objective is to minimize both the expected regret and the expected (signed) cumulative constraint violation, defined as
$\mathsf{CCV}_T = \mathbb{E}\left[\sum_{t=1}^T g_t(x_t)\right]^+.$
Using a similar Lyapunov-based approach with a quadratic Lyapunov function (see Appendix~\ref{sec:ftpl_pf} for details), we reduce the above constrained problem to an unconstrained OLO problem with modified cost functions. Specifically, at round $t$, we define the linear surrogate cost function 
\[
\hat{f}_t(x) = \langle V f_t + 2 Q(t-1)\, g_t,\; x \rangle, ~~ t \geq 1,
\]
where $V > 0$ is a parameter to be specified later. Applying the $\mathsf{FTPL}$ algorithm with Gaussian perturbations to the sequence $\{\hat{f}_t\}_{t=1}^T$ yields the following result.

\begin{theorem}[Projection-Free OLO with Adversarial Constraints]
\label{ftpl_perf_thm}

Consider the above $\mathsf{FTPL}$-based policy in the full-information setting with $V = \sqrt{T}.$ It achieves the following bounds on the expected regret and expected cumulative constraint violation:
\begin{align*}
\mathbb{E} \mathsf{Regret}_T  \leq \mathcal{O}(T^{3/4}), 
~~\mathbb{E}\mathsf{CCV}_T \leq \mathcal{O}(T^{3/4}).
\end{align*}
\end{theorem}

The proof has been deferred to Appendix \ref{sec:ftpl_pf}.

\section{Numerical Simulations}
To show the practical benefit of our algorithm, we demonstrate the efficacy of our algorithm in the problem of  Online Shortest Path with Delays. We have detailed the experiments in section \ref{expts}. We compare our algorithm against three other baselines --- the projection-based algorithm proposed by \cite{sinha2024optimal} and the projection-free algorithms of \cite{garber2024projectionfree} and \cite{wang2025revisitingprojectionfreeonlinelearning}. The \cite{sinha2024optimal} algorithm has better regret and violation performance but orders of magnitude worse computational efficiency. Among projection-free algorithms, ours has the best regret and violation performance. This is theoretically supported by the sub-optimal regret and violation bounds of \cite{garber2024projectionfree} and the use of the inefficient doubling trick in the case of \cite{wang2025revisitingprojectionfreeonlinelearning}.
\section{Conclusion} \label{conclusion}
%It will be interesting to see whether one can achieve better regret bound by extending the projection-free algorithm proposed by \citet{mhammedi2022efficient}. 
%In this paper, we proposed an efficient projection-free OCO policy with time-varying adversarial constraints 
This paper introduces an efficient projection-free policy for OCO problems with time-varying adversarial constraints
for both full-information and bandit feedback settings. The proposed algorithm uses a new, Lipschitz-adaptive version of the Online Condition Gradient algorithm and achieves $\tilde{O}(T^{\nicefrac{3}{4}})$ bound for both Regret and CCV metrics in the full information setting, thus improving upon the state-of-the-art results. Additionally, for constrained online linear optimization over combinatorial structures, we present an efficient Follow-the-Perturbed-Leader (FTPL) policy that directly outputs discrete feasible actions without requiring expensive post-hoc sampling, while achieving slightly improved Regret and CCV bounds of $O(T^{3/4})$. In the bandit setting, we use an OCG-type algorithm to obtain the same bounds of $\tilde{O}(T^{\nicefrac{3}{4}})$ on expected regret and violation, and the analysis offers novel perspectives. 
\section{Acknowledgement}
The work of Abhishek Sinha was supported by the Department of Atomic Energy, Government of India, under project no.\ RTI4014 and by a Google India faculty research award. Palash Dey thanks the Anusandhan National Research Foundation (ANRF) (erstwhile Science, Education, and Research Board (SERB)), Government of India, for supporting this work through Core Research Grant under file no. CRG/2022/003294.
%\newpage
%\clearpage
\bibliography{OCO}
\bibliographystyle{plainnat}
\clearpage
\newpage
\appendix
\onecolumn
\textbf{\Large Appendix}
\section{Related Work} \label{related}
\paragraph{Unconstrained OCO:}
In a pioneering work, \citet{zinkevich2003online} demonstrated that the well-known projected online gradient descent (OGD) strategy achieves an $O(\sqrt{T})$ regret for OCO with bounded sub-gradients. Subsequently, various studies have introduced adaptive and parameter-free variations of OGD \citep{hazan2007adaptive, orabona2018scale}. Textbook discussions on the OCO framework and related algorithms can be found in \citet{orabona2019modern, hazan2022introduction}. 
\begin{table*}[h]
%\vskip 0.15in
%\begin{center}
%\begin{small}
%\begin{sc}
\begin{tabular}{llllll} %lccccr
\toprule
Reference & Regret & CCV & Complexity per round & Assumptions&\\
\midrule
\citet{mahdavi2012trading} & $O(\sqrt{T})$& $O(T^{\nicefrac{3}{4}})$& Proj.& Fixed Constraints \\\citet{jenatton2016adaptive}& $O(T^{\max(\beta, 1-\beta) })$& $O(T^{1- \beta/2})$& Proj& Fixed Constraints& \\
\citet{yu2017online}    & $O(\sqrt{T})$& $O(\sqrt{T})$& Proj& Slater \& Stochastic Constr. &       \\
\citet{neely2017online}     & $O(\sqrt{T})$ & $O(\sqrt{T})$& Proj.& Slater&\\
\citet{yu2020low} & $O(\sqrt{T})$ & $O(1)$& Conv-OPT & Slater \& Fixed Constraints&\\
\citet{yi2021regret}     & $O(\sqrt{T})$& $O(T^{\nicefrac{1}{4}})$& Conv-OPT& Fixed Constraints&\\
\citet{guo2022online}    & $O(\sqrt{T})$& $O(T^{\nicefrac{3}{4}})$& Conv-OPT& ---& \\
\citet{yi2023distributed}    & $O(T^{\max(\beta,1-\beta)})$& $O(T^{1-\beta})$& Conv-OPT& Slater &\\
%\citet{sinha2023playing}    & $O(\sqrt{T})$& $O(T^{\nicefrac{3}{4}})$& Proj& ---&        \\
%\citet{sinha2023playing}    & $O(\sqrt{T})$& $O(\sqrt{T})$& Proj& $\textrm{Regret}_T \geq 0$ &       \\
\citet{sinha2024optimal}   & $O(\sqrt{T})$& $\tilde{O}(\sqrt{T})$& Proj& --- &\\
\citet{garber2024projectionfree}   & $O(T^{\nicefrac{3}{4}})$& $O(T^{\nicefrac{7}{8}})$& LP& --- &\\
\citet{garber2024projectionfree}   & $O(T^{\nicefrac{3}{4}})$& $O(T^{\nicefrac{7}{8}})$& Conv-OPT& --- &\\
\citet{wang2025revisitingprojectionfreeonlinelearning}   & $O(T^{\nicefrac{3}{4}})$& $\tilde{O}(T^{\nicefrac{3}{4}})$& LP& Use of doubling trick &\\
\textbf{This paper}   & $\tilde{O}(T^{\nicefrac{3}{4}})$& $\tilde{O}(T^{\nicefrac{3}{4}})$ & \textbf{LP} & \textbf{---} &\\
\bottomrule
\end{tabular}
%\end{sc}
%\end{small}
%\end{center}
%\vskip -0.1in
\vspace{2 mm}
\caption{Summary of the results for the COCO problem with convex cost and convex constraint functions. The parameter $\beta \in [0,1]$ is any fixed constant. Abbreviations have the following meanings- \textsc{Proj}: Euclidean projection on the decision set, \textsc{Conv-OPT}: Solving a convex optimization problem over the decision set, \textsc{LP}: solving a linear optimization problem over the decision set.}
\label{comp-table}
\end{table*}

{\bf Constrained OCO (COCO): 

 Time-invariant constraints:} A number of papers have explored COCO with time-invariant constraints, where $g_{t,i} = g_i$ for all $t \geq 1$ \citep{yuan2018online, jenatton2016adaptive, mahdavi2012trading, yi2021regret}. These studies typically assume that the constraint functions are known beforehand. However, they permit the policy to be infeasible in any round to circumvent the expensive projection step involved in traditional projected OGD policy. The primary aim of this line of work is to design a computationally efficient policy by avoiding the projection step while maintaining small regret and cumulative constraint violation bounds. 

{\bf Time-varying constraints:} Tackling the COCO problem with time-varying constraint functions presents greater difficulties. Aside from \citet{neely2017online} and \citet{georgios-cautious}, most existing work involves constructing a Lagrangian function and subsequently updating primal and dual variables in an online fashion \citep{yu2017online, pmlr-v70-sun17a, yi2023distributed}. However, the resulting algorithms turn out to be inefficient with sub-optimal regret/CCV bounds. Both \citet{neely2017online} and \cite{georgios-cautious} employ the drift-plus-penalty (DPP) methodology to address the constrained problem under additional assumptions. Specifically, \citet{neely2017online} proposed a policy grounded on DPP for COCO with Slater's condition, which assumes strict feasibility in the sense that $g_{t}(x^\star) < -\eta$ for some $\eta>0$ $\forall t$. This clearly precludes standard constraint functions without slack (such as functions of the form $\max(0, g_t(x))$). Additionally, the bounds derived under Slater's condition depend inversely on the Slater constant $\eta$ (generally hidden under the big-Oh notation). As $\eta$ may be arbitrarily small, these bounds might be excessively loose. \citet{georgios-cautious} augmented \citet{neely2017online}'s work by incorporating a less stringent feasibility assumption without Slater's condition.  In more recent work, \citet{sinha2024optimal} presented a simple COCO algorithm which uses a Euclidean projection step and achieves state-of-the-art performance guarantees. \citet{sarkar2025online} extended their work by considering a class of non-convex functions with a strengthened benchmark. \citet{sarkar2025optimal} also came up with an anytime version of their policy by making innovative use of time-varying Lyapunov functions. They further derived an adaptive bound on the CCV. 

{\bf Projection-free OCO and COCO:}

%To make these algorithms viable for practical applications, attention must be paid not only to their performance but also to their computational efficiency. 
For these algorithms to be practical, both performance and computational efficiency must be considered.
It is well-established that the projection step commonly employed in OCO and COCO algorithms in every round is highly computationally expensive. This has motivated the development of the so-called \emph{projection-free} algorithms that replace the expensive projection step with a linear optimization step.

One of the first works that studies this problem in the OCO setup is \citet{hazan2012projection}. Further contributions to the area of projection-free OCO have been made by \citet{garber2023newprojectionfreealgorithmsonline,garber2021efficient}. There also exist projection-free works that, instead of an LP step, assume some strong oracle, such as the access to a membership oracle in \citet{garber2022new} and \citet{mhammedi2022efficient}. While these works achieve strong regret guarantees, the overall algorithm still might be computationally expensive outside of specific use cases. 

It is still an open question as to what the asymptotic lower bounds of Frank-Wolfe based projection-free methods are. Very recently, \cite{weibel2025optimizedprojectionfreealgorithmsonline} provide strong numerical evidence that the lower bound on regret for these methods is $\Omega(T^{3/4})$. However, a formal proof for the same remains elusive.

%For example, for solving an LP, one needs to use the membership oracle with an algorithm like Ellipsoid which has high computational complexity. This defeats the very purpose of using projection-free algorithms as projections would typically be faster to compute.

\iffalse
As mentioned previously, there have been a few papers that have proposed projection-free methods in the \textit{restricted} COCO setting. We are referring to those who assume time-invariant constraints and have prior knowledge of them. 
\fi
\citet{garber2024projectionfree} was the first to address the design of projection-free algorithms for general COCO.
%is the first work which deals with the problem of designing projection-free algorithms for general COCO. 
This paper also makes use of a drift-plus-penalty technique like other COCO works, except that it does not make use of a projection step. It uses the online Frank-Wolfe algorithm to iteratively move towards feasibility without explicit projection. It must be noted, however, that while the algorithm that they have proposed accesses the admissible set only via call to LP solvers, at one point, it has to do a convex optimization step over an Euclidean ball. While this is not the same as performing convex optimization over the admissible set, this is still a costly operation compared to linear optimization. They further propose a second algorithm where they perform online primal-dual (sub)gradient
updates w.r.t. the dual-regularized Lagrangian functions. This algorithm does not make use of convex optimization, but the downside is that the regret and violation bounds do not hold for a subsequence of cost and constraint functions. Instead, they only hold when the entire sequence is considered at once. Both algorithms divide the horizon into disjoint blocks and only perform updates at the end of a block. They also extend their results into the bandit setting by employing the standard reduction using smoothed functions.
\subsection{Relation to \texorpdfstring{\citet{wang2025revisitingprojectionfreeonlinelearning}}{Wang et al. 2025}}
\label{wang_comparison}
We dedicate this subsection to specifically situating our contributions with respect to concurrent work by \citet{wang2025revisitingprojectionfreeonlinelearning} regarding the OCG-based policy. At first glance, it would appear that our guarantees in the full-information and bandit settings are overlapping with theirs. However, we now argue, despite seemingly overlapping guarantees, our approach is fundamentally different from theirs and how we make a unique contribution to the literature. 

At the outset, we must mention the similarity between our paper and \citet{wang2025revisitingprojectionfreeonlinelearning}. At a high level, both papers minimize regret and violation by constructing a special surrogate loss $\Tilde{f}_t$ by additively combining the cost $f_t$ and constraint function $g_t$ in a proportion which depends on the violation $Q(t)$ till time $t$. After that, the surrogate loss is passed to the Online Conditional Gradient (OCG) algorithm. A major technical limitation of existing approaches to the analysis of the OCG algorithm is that they consider a uniform Lipchitz constant $G$ for all cost functions up to time $T.$ However, in our setting, since $Q(t)$ at round $t$ is unknown, we cannot assign a Lipschitz constant to the surrogate cost $\Tilde{f}_t$ at time $t.$ This is the main technical challenge --- to design an \textit{adaptive} OCG policy where we don't need to know a uniform Lipchitz constant across all time steps. How we achieve this is a point of departure from \citet{wang2025revisitingprojectionfreeonlinelearning}.
\begin{enumerate}
    \item \textbf{Critique of the Doubling Trick Approach in Wang et al. (2025):} In the full information setting, \citet{wang2025revisitingprojectionfreeonlinelearning} makes use of a Franke-Wolfe type policy like us. However, to handle the unknown gradient norms required for setting the learning rate, they rely on a doubling trick. The doubling trick has not been viewed very favorably by the community. See the discussion in \citet{zhang2024improving}, \citet{kwon2014continuous}, \citet{luo2014towards} and \citet{besson2018doubling} for more details. These works have considered the doubling trick to be aesthetically inelegant, intuitively wasteful and practically suboptimal (as confirmed by our experiments) as it involves breaking up the time horizon into phases and then restarting the algorithm from scratch at the beginning of every phase. 
    
    It is to be noted that while the typical use of the doubling trick is in estimating the length of the time horizon, which is a priori unknown, it is a general technique and can be used even when some other quantity is unknown. In \citet{wang2025revisitingprojectionfreeonlinelearning} it was used because the value of $\tilde{G}$, the Lipchitz constant which upper-bounds the gradient of all surrogate losses across timesteps was unknown. To do so, they estimated a value of $\tilde{G}$ and ran the algorithm till it was valid, once it was not, they doubled it and started the algorithm from scratch which means that it loses all the accumulated knowledge upto that point. This is intuitively wasteful because the foundational idea behind OCO is to improve our predictions by accumulating knowledge. By periodically throwing away the accumulated information, we are going against this intuition. This type of restarting technique, the doubling trick, is often considered a last resort and brute-force approach when all other adaptive strategies have failed, which is why it is also considered aesthetically inelegant, as its use provides limited inspiration to the community. In our paper, we come up with a more elegant adaptive strategy that removes the need to use the doubling trick. This makes our analysis both more novel and more impactful in the context of the online learning community. As we show in our experiments (see section \ref{expts}), the doubling trick is also practically suboptimal, underperforming our algorithm despite having similar theoretical guarantees.
    
    \item \textbf{Our Intrinsically Adaptive Method:} We sidestep the use of the doubling trick by adapting the Frank-Wolfe algorithm to our setting in a more elegant manner. We too face the technical challenge of adapting the learning rate to the unknown and time-varying Lipschitz constants of the surrogate cost functions. In contrast to restarting, our Algorithm \ref{ocg_alg} introduces a novel, intrinsically adaptive version of the Online Conditional Gradient (OCG) method. This algorithm adjusts its learning rate at every single step based on the most recently observed gradient norm. This approach avoids the disruptive restarts of the doubling trick, resulting in a more stable algorithm. Furthermore, in developing this solution, we derive the first known scale-free adaptive regret bounds for the OCG policy, a theoretical result of independent interest to the online learning community.

In summary, while \citet{wang2025revisitingprojectionfreeonlinelearning} divides the time-horizon into chunks and then runs the usual OCG in those chunks, our method differs more fundamentally from the usual OCG, as the Lipchitz-constants are taken so that they can be time-varying at every timestep. We don't reduce our \textit{adaptive} OCG into multiple sequential OCGs started from scratch and instead consider one contiguous OCG-like algorithm with time-varying Lipchitz constants. It takes considerable technical novelty on our part to derive scale-free regret bounds for this adaptive algorithm, which we then use in our analysis. 

Next, we talk about the bandit section in \citet{wang2025revisitingprojectionfreeonlinelearning} and how the analysis is flawed and the results don't actually hold up. Thus, our guarantees in the bandit setting are the first of their kind. 

\item \textbf{The erroneous assumption of non-stochasticity of $Q(t)$ in the bandit analysis of Wang et al. (2025):} It is well-known that in the bandit convex optimization setting, since we need to add a random perturbation to our actions, we do not bound our regret but the expected regret as the regret becomes as inherently stochastic quantity in that case. Due to similar reasoning $Q(t)$ is also an inherently stochastic quantity. Indeed, both \cite{garber2024projectionfree} and our paper bound the expectation of $Q(t)$ instead of $Q(t)$ for this very reason. On the other hand, \citet{wang2025revisitingprojectionfreeonlinelearning} proceeds with their bandit analysis as if $Q(t)$ were a non-stochastic quantity. We are not being pedantic here --- this is a fundamental flaw and renders their analysis invalid as the analysis would be non-trivially different if they were to deal with $\mathbb{E}[Q(t)]$ in their analysis.

\item \textbf{On the form of the surrogate loss and its unique significance in the bandit setting:} Notwithstanding the above point, there are other, more subtle flaws in the bandit analysis of \citet{wang2025revisitingprojectionfreeonlinelearning}. In that paper, they effectively reused the form of the surrogate loss used in the full-information setting in the bandit setting as well. However, as we point out in our bandit section (section \ref{bandit_sec}), this is not acceptable, and we instead have to come up with a new surrogate loss $\Tilde{f}_t$ that depends on $Q(t-1)$ instead of $Q(t).$ Put briefly, this is because making $Q(t)$ part of the loss implies that we are making the function itself dependent on the random action which is problematic for unbiased gradient estimation, as the only source of randomness during gradient estimation should be the random perturbation introduced at time $t$ and the function itself should be deterministic given all the randomness of previous rounds. Thus, under the surrogate loss assumed in \citet{wang2025revisitingprojectionfreeonlinelearning}, bandit learning is impossible. These minutiae are important and must not be abstracted out in a superficial analysis. This is a more subtle point and easier to miss, and thus, this is also an independent contribution of ours to the field of constrained BCO.
\end{enumerate}
\section{Proofs for Projection-Free COCO}
\subsection{Proof of Theorem \ref{ocg-reg-thm}}   
Our proof combines the arguments in \citet[Theorem 23]{hazan2022introduction} with the adaptive regret bound of the FTRL policy with a time-varying Euclidean regularizer \citep[Corollary 7.9]{orabona2019modern}. For reference, the FTRL policy with a time-varying Euclidean regularizer is described in Algorithm \ref{ftrl_alg}.
Since \citet[Theorem 23]{hazan2022introduction} considers time-invariant non-adaptive regularizers, non-trivial extensions are required to derive an adaptive regret bound for the OCG policy. We first consider the following fictitious action sequence: 
\begin{eqnarray} \label{ftrl-actions}
    x^\star_t = \arg \min\limits_{x \in \mathcal{X}} F_{t-1}(x),~~t \geq 1,
    \end{eqnarray}
where the regularized cumulative loss function $F_t(\cdot)$ is defined in Eqn.\ \eqref{surr_fn}. Recall that $\{x_t\}_{t\geq 1}$ is the action sequence generated by the OCG algorithm. 
On the other hand, it is easy to see that the sequence $\{x^\star_t\}_{t\geq 1}$ corresponds to the iterates of the Follow-the-Regularized-Leader (FTRL) algorithm \citep[Algorithm 13]{hazan2022introduction} when run with the auxiliary cost function sequence $\{\tilde{f}_t\}_{t\geq 1},$ where the function $\tilde{f}_t$ is a shifted version of the original cost function $\hat{f}_t$: 
\begin{eqnarray} \label{f_tilde}
	\tilde{f}_t(x) := \hat{f}_t(x +  x_t - x^\star_t), ~~t \geq 1.
\end{eqnarray}
%Please refer to Algorithm \ref{ftrl_alg} described subsequently.
This can be verified by referring to the FTRL algorithm, described in Algorithm \ref{ftrl_alg},
along with the fact that $\nabla \tilde{f}_t(x_t^\star) = \nabla \hat{f}_t(x_t) \equiv \nabla_t.$ Clearly, the function $\tilde{f}_t$ is also $L_t$-Lipschitz.
%in FTRL the gradients we use are the gradient of the cost functions evaluated at $x^*_t$s, whereas in OCG, the gradients are evaluated at $x_t$. 
%Furthermore, we have $\tilde{f}_t(x_t^\star) = \hat{f}_t(x_t)$.
We emphasize that the action sequence $\{x^\star_t\}_{t\geq 1}$ is not necessarily played by the algorithm and is introduced here purely for the sake of analysis. 
The regret with respect to any fixed benchmark action $x^\star \in \mathcal{X}$ can be decomposed into the sum of two terms, $(A)$ and $(B)$, as described below:
\begin{eqnarray} \label{regret_bd}
  && \textrm{Regret}_T(\OCG) 
   =\sum\limits_{t=1}^{T} \big(\hat{f}_t(x_t) - \hat{f}_t(x^\star)\big)  
    = \sum\limits_{t=1}^{T} \big(\hat{f}_t(x_t) - \hat{f}_t(x^\star_t)\big) + \sum\limits_{t=1}^{T} \big(\hat{f}_t(x^\star_t) - \hat{f}_t(x^\star)\big) \nonumber\\
    &\stackrel{(a)}{\leq} & \underbrace{\sum_{t=1}^T L_t||x_t-x_t^\star||}_{(A)}+
    \underbrace{\sum\limits_{t=1}^{T} \big(\hat{f}_t(x^\star_t) - \hat{f}_t(x^\star)\big)}_{(B)},
  % &\stackrel{(b)}{\leq} &   \underbrace{\sum_{t=1}^T L_t||x_t-x_{t+1}^\star||}_{(A)} + \underbrace{\sum_{t=1}^T L_t||x_{t+1}^\star-x_{t}^\star||}_{(B)} \nonumber \\
  % && +\underbrace{\sum\limits_{t=1}^{T} \big(\hat{f}_t(x^\star_t) - \hat{f}_t(x^\star)\big)}_{(C)}
\end{eqnarray}
where inequality $(a)$ follows from the Lipschitz continuity of the functions $\{\hat{f}_t, t \geq 1\}$. We now bound the above two terms separately.
%and inequality (b) follows from the Triangle inequality.
\paragraph{Bounding term $(A)$:} 
 Using triangle inequality, we can further upper-bound term $(A)$ with the sum of two terms, $(C)$ and $(D)$, as follows:
 \begin{eqnarray} \label{term-c-d}
 	\underbrace{\sum_{t=1}^T L_t||x_t-x_t^\star||}_{(A)} &\leq& \underbrace{\sum_{t=1}^T L_t||x_t-x_{t+1}^\star||}_{(C)} 
 	+ \underbrace{\sum_{t=1}^T L_t||x_{t+1}^\star-x_{t}^\star||}_{(D)}.
 \end{eqnarray}
%\paragraph{Bounding term (a):} 
%Lemma \ref{term-a-lemma} gives an upper bound to term $(C)$. 
The following two lemmas give upper bounds to term $(C)$ and term $(D)$, respectively. 
\begin{lemma} \label{term-a-lemma}
	Term (C) in Eqn.\ \eqref{term-c-d} can be upper bounded as
	\begin{eqnarray*}
		(C)\equiv \sum_{t=1}^T L_t||x_t-x_{t+1}^\star|| \leq 12 D L_T T^{3/4} \sqrt{\log(Te L_T/L_1)}).
	\end{eqnarray*}
\end{lemma}
See Appendix \ref{term-a-lemma-pf} for the proof of Lemma \ref{term-a-lemma}.

\begin{lemma} \label{term-d-lemma}
	Term (D) in Eqn.\ \eqref{term-c-d} can be upper bounded as
	\begin{eqnarray*}
		(D)\equiv \sum_{t=1}^T L_t||x_{t+1}^\star-x_{t}^\star|| \leq 3D L_T \sqrt{T}.
	\end{eqnarray*}
\end{lemma}
The proof of the above result follows from the stability property of the FTRL policy. See Appendix \ref{term-d-lemma-pf} for the proof of Lemma \ref{term-d-lemma}. Lemma \eqref{term-a-lemma} and \eqref{term-d-lemma}, taken together, upper bounds Term $(A)$ in Eqn.\ \eqref{regret_bd}. 

\paragraph{Bounding term (B):}
The $t$\textsuperscript{th} term inside the summation of term (B) in Eqn.\ \eqref{regret_bd} can be upper bounded as follows: 
\begin{eqnarray*}
	\hat{f}_t(x_t^\star) - \hat{f}_t(x^\star)\stackrel{(c)}{=}\tilde{f}_t(x_t^\star -x_t+x_t^\star) - \tilde{f}_t(x^\star - x_t+x_t^\star) 
	\stackrel{(d)}{\leq} \tilde{f}_t(x_t^\star) - \tilde{f}_t(x^\star)  + 2L_t||x_t^\star-x_t||, 
\end{eqnarray*}
where (c) follows from the Definition \eqref{f_tilde} and (d) follows from the fact that the auxiliary cost function $\tilde{f}_t$ is $L_t$-Lipschitz. Hence, term $(B)$ can be upper bounded as:
\begin{eqnarray} \label{intermediate_bd}
	(B) \leq  \textrm{Regret}_T(\textrm{FTRL})+ 2(A),
\end{eqnarray}
where $\textrm{Regret}_T(\textrm{FTRL}) \equiv \sum_{t=1}^T \big(\tilde{f}_t(x_t^\star) - \tilde{f}_t(x^\star)\big) $ is the regret of the FTRL policy for the auxiliary cost function sequence $\{\tilde{f}_t\}_{t=1}^T$ with adaptive learning rates. The pseudocode of the FTRL policy is given in Algorithm \ref{ftrl_alg} for reference \citep{orabona2019modern}. 
\begin{algorithm}
\caption{Follow-the-Regularized-Leader}\label{ftrl_alg}
\begin{algorithmic}[1]
\STATE \textbf{Inputs:} Convex decision set $\mathcal{X},$ sequence of convex cost functions $\{l_t: \mathcal{X} \to \mathbb{R}\}_{t \geq 1},$ $\textrm{diam}(\mathcal{X}) = D.$
\STATE \textbf{Parameters:} Closed and $2$-strongly convex function (w.r.t. the Euclidean norm)  $\psi(x) \equiv ||x-x_1||^2$, adaptive regularizer sequence $\psi_t(x)= \frac{\psi(x)}{\eta_{t-1}} $, where $\eta_{t+1} \leq \eta_t, \forall t \geq 1$.
\STATE Initialize $x_1 \in \mathcal{X}$ arbitrarily 
\FOR{$t=1:T$}
\STATE Play $x_t,$ observe $l_t(\cdot),$ and compute $\nabla_t \equiv \nabla l_t(x_t).$ 
\STATE
Compute \[x_{t+1}= \arg \min_{x \in \mathcal{X}}  \sum_{\tau=1}^{t} \nabla_\tau^{\top} x+\frac{||x-x_1||^2}{\eta_t}. \]
\ENDFOR
\end{algorithmic}
\end{algorithm}
The following result gives an upper bound to the FTRL policy with adaptive learning rates. 
\begin{theorem}(\citet[Corollary 7.9]{orabona2019modern})\label{ftrl_bound}
   The FTRL policy, described in Algorithm \ref{ftrl_alg}, achieves the following regret bound for any sequence of convex cost functions $\{l_t\}_{t \geq 1}$ and for any feasible action $u \in \mathcal{X}:$
    \[
    \sum\limits_{t=1}^{T}l_t(x_t) - \sum\limits_{t=1}^{T}l_t(u)
    \leq
    \frac{D^2}{\eta_{T}} + \frac{1}{4} \sum\limits_{t=1}^{T}\eta_{t-1} ||\nabla_t||^2. 
    \]
   % For all $\nabla_t \in \partial l_t(x_t)$.
\end{theorem}

%In our case, we have $\psi(x) = ||x - x_1||^2_2$ as the regularizer. This function is 1-strongly convex w.r.t. the $L_2$-norm. So $||\nabla_t||_* = ||\nabla_t||$ and $\mu = 1$. Hence, the above result implies that 
Setting $l_t=\tilde{f}_t,$ we have the following regret bound in our case:
\begin{eqnarray} \label{ftrl-regret-bd}
	&&\textrm{Regret}_T(\textrm{FTRL}) 
	\leq \frac{D^2}{\eta_{T}} + \frac{1}{4} \sum\limits_{t=1}^{T}\eta_{t-1} L_t^2 
	\leq  2DL_T T^{\frac{3}{4}} + \frac{D}{8T^{3/4}}\sum_{t=1}^T \frac{L_t^2}{L_{t-1}}\nonumber \\
	&\leq & 2DL_T T^{\frac{3}{4}} + \frac{D}{8T^{3/4}}\big(\max_{1\leq t \leq T} \frac{L_{t}}{L_{t-1}}\big) \sum_{t=1}^T L_t \leq 2DL_T T^{\frac{3}{4}} + \frac{DL_TT^{1/4}}{8}\big(\max_{1\leq t \leq T} \frac{L_{t}}{L_{t-1}}\big).
\end{eqnarray} 
The final result now follows upon combining Eqn.\ \eqref{regret_bd} with Lemma \ref{term-a-lemma}, Eqn.\ \eqref{intermediate_bd}, and the FTRL regret bound \eqref{ftrl-regret-bd}, yielding 
\begin{eqnarray*}
	\textrm{Regret}_T(\textrm{OCG}) \leq 3(A) +  \textrm{Regret}_T(\textrm{FTRL})
	\leq  47 D L_T  T^{3/4} \sqrt{\log(Te L_T/L_1)})\bigg(1+ \frac{1}{376\sqrt{T}}(\max_{1\leq t \leq T} \frac{L_{t}}{L_{t-1}})\bigg). 
\end{eqnarray*}
\subsection{Supporting Lemmata}
In the analysis of Section \ref{analysis_sec}, we state the following two supporting Lemmas, which were used in the argument there.
\begin{lemma} \label{ratio-bd}
Choosing $\Phi(x)=x^p$ with $p =\log T$ and for any $V>0,$ we have 
	\begin{eqnarray*}
		\max_{1\leq t \leq T} \frac{L_{t}}{L_{t-1}} \leq 1+e.
	\end{eqnarray*}
\end{lemma}
	See Section \ref{ratio-bd-proof} for the proof of this Lemma.

	\begin{lemma}
    \label{sqrt-ratio}
   Choosing $\Phi(x)=x^p$ with $p =\log T$ and for any $V>0,$ we have 
	\begin{eqnarray*}
		\sqrt {\log \frac{L_{T}}{L_{1}}} \leq \log T.
	\end{eqnarray*}
\end{lemma}
See Section \ref{sqrt-ratio-proof} for the proof.
\subsection{Proof of Lemma \ref{term-a-lemma}} \label{term-a-lemma-pf}
%\cmt{change to $x^*_{t+1} \to x^*_{t+1}$}.
Recall the definition of the regularized cumulative cost function from Eqn.\ \eqref{surr_fn}:
\begin{eqnarray} \label{reg-cost}
F_t(x) = \sum_{\tau=1}^{t} \nabla_{\tau}^Tx + \frac{||x - x_1||^2}{\eta_t},	
\end{eqnarray}
with the adaptive learning rate sequence $\{\eta_t\}_{t \geq 1}$ given in Algorithm \ref{ocg_alg}.
Clearly, the quadratic function 
$F_t$ \text{is} $\frac{2}{\eta_t}$-\text{smooth} and $\frac{2}{\eta_t}$-\text{strongly convex} w.r.t. the standard Euclidean norm. Similar to the proof in \citep[Theorem 23]{hazan2022introduction}, we define \[h_t(x) \stackrel{(\textrm{def})}{=} F_t(x) - F_t(x^\star_{t+1}), ~~~h_t \stackrel{(\textrm{def})}{=} h_t(x_t),\] where $x^\star_{t+1} \equiv \arg\min_{x \in \mathcal{X}} F_{t}(x),$ as defined in Eqn.\ \eqref{ftrl-actions}. Now recall that the OCG policy chooses the next action as $x_{t+1} = (1-\sigma_t)x_t + \sigma_t v_t$.
Hence,
\begin{eqnarray*}
	h_t(x_{t+1}) &=& F_t(x_{t+1}) - F_t(x_{t+1}^\star)\\
	&=& F_t(x_t + \sigma_t (v_t-x_t)) - F_t(x_{t+1}^\star)  \\
	&\stackrel{(a)}{\leq}& F_t(x_t)-F_t(x_{t+1}^\star)+ \sigma_t\langle \nabla F_t(x_t), v_t-x_t\rangle + \frac{\sigma_t^2 ||v_t-x_t||^2}{\eta_t}  \\
	&\stackrel{(b)}{\leq}& F_t(x_t)-F_t(x_{t+1}^\star)+ \sigma_t\langle \nabla F_t(x_t), x_{t+1}^\star-x_t\rangle + \frac{\sigma_t^2 D^2}{\eta_t}  \\
	&\stackrel{(c)}{\leq}& (1-\sigma_t)\big(F_t(x_t)-F_t(x_{t+1}^\star)\big) + \frac{\sigma_t^2 D^2}{\eta_t},
\end{eqnarray*}
where (a) uses the smoothness property of $F_t,$ (b) follows from the definition of $v
_t$ as given in Eqn.\ \eqref{v-def}, and (c) follows from the convexity of $F_t.$ Thus, the above yields
\begin{eqnarray} \label{eq1}
h_{t}(x_{t+1}) \leq (1-\sigma_t)h_t + \frac{\sigma_t^2 D^2}{\eta_t}.
\end{eqnarray}
Furthermore, we have
\begin{eqnarray} \label{h-bd}
h_{t+1} 
&=& F_{t+1}(x_{t+1}) - F_{t+1}(x_{t+2}^\star) \nonumber \\
&=& F_t(x_{t+1}) + \nabla_{t+1}^{T}x_{t+1} + ||x_{t+1} - x_1||^2\big(\frac{1}{\eta_{t+1}} - \frac{1}{\eta_t}\big) - F_t(x_{t+2}^\star)  - \nabla_{t+1}^{T}x_{t+2}^\star - ||x_{t+2}^\star - x_1||^2\big(\frac{1}{\eta_{t+1}} - \frac{1}{\eta_t}\big)\nonumber\\
&\leq& h_{t}(x_{t+1}) + L_{t+1}||x_{t+1}-x^\star_{t+2}|| + D^2\big(\frac{1}{\eta_{t+1}} - \frac{1}{\eta_t}\big),
\end{eqnarray}
where in the last step, we have used Cauchy-Schwarz inequality along with the fact that $||\nabla_t|| \leq L_t.$
Since, $F_{t+1}$ is $\frac{2}{\eta_{t+1}}$-strongly convex, and $x^\star_{t+2} \equiv \arg\min_{x \in \mathcal{X}} F_{t+1}(x),$ we have
\begin{eqnarray*}
h_{t+1}= h_{t+1}(x_{t+1}) = F_{t+1}(x_{t+1}) - F_{t+1}(x_{t+2}^\star) \geq \frac{1}{\eta_{t+1}} ||x_{t+1} - x_{t+2}^\star||^2.
\end{eqnarray*}
This implies that 
\begin{eqnarray} \label{u-t-bd}
 ||x_{t+1} - x_{t+2}^\star|| \leq \sqrt{\eta_{t+1}h_{t+1}}.
\end{eqnarray}
By substituting this bound into Eqn.\ \eqref{h-bd}, we conclude that  
\[
h_{t+1} \leq h_{t}(x_{t+1}) + L_{t+1}\sqrt{\eta_{t+1}h_{t+1}} + D^2\big(\frac{1}{\eta_{t+1}} - \frac{1}{\eta_t}\big).
\]
Substituting the upper bound from Eqn.\ \eqref{eq1} into the above, we obtain
\[
h_{t+1} \leq (1-\sigma_t)h_t + \frac{\sigma_t^2 D^2}{\eta_t} + L_{t+1}\sqrt{\eta_{t+1}h_{t+1}} + D^2(\frac{1}{\eta_{t+1}} - \frac{1}{\eta_t}).
\]
Multiplying both sides of the above equation by $\eta_{t+1},$ we obtain
%\subsection*{First method}
%\[
%x^2 \leq bx + c \implies x^2 \leq b^2 + 2c
%\]
%\[
%h_{t+1} \leq 2(1-\sigma_t)h_t + \frac{D^2}{\eta_t}\sigma_t^2 +  D^2(\frac{1}{\eta_{t+1}} - \frac{1}{\eta_t}) + L_t^2\eta_{t+1}
%\]
%\subsection*{Second method}
\begin{eqnarray*}
\eta_{t+1}h_{t+1} &\leq& (1-\sigma_t)\eta_{t+1}h_t + \eta_{t+1}\frac{\sigma_t^2D^2}{\eta_t} + \eta_{t+1}L_{t+1}\sqrt{\eta_{t+1}h_{t+1}} + D^2(1 - \frac{\eta_{t+1}}{\eta_t}) \\
&\leq& (1-\sigma_t)\eta_{t}h_t + \sigma_t^2 D^2+ \eta_{t+1}L_{t+1}\sqrt{\eta_{t+1}h_{t+1}} + D^2(1 - \frac{\eta_{t+1}}{\eta_t}),
\end{eqnarray*}
where in the last step, we have used the fact that $\eta_t \geq \eta_{t+1}, \forall t\geq 1.$ Let us now  define the sequence $u_t \equiv \eta_t h_t, t\geq 1.$ From the above, we have
\begin{eqnarray} \label{term-d}
u_{t+1} \leq (1-\sigma_t)u_t + \sigma_t^2 D^2 + \underbrace{\eta_{t+1}L_{t+1}\sqrt{u_{t+1}}}_\text{\clap{(c)}} + \underbrace{D^2(1 - \frac{\eta_{t+1}}{\eta_t})}_\text{\clap{(d)}}.
\end{eqnarray}
Now, term (c) can be upper bounded as follows: 
\begin{eqnarray*}
\eta_{t+1} L_{t+1} \sqrt{u_{t+1}}&=&(\sqrt{D} \eta_{t+1} L_{t+1})^{2 / 3}\left(\frac{\eta_{t+1} L_{t+1}}{D}\right)^{1 / 3} \sqrt{u_{t+1}} \\
&\stackrel{(e)}{\leq}& \frac{1}{2}(\sqrt{D} \eta_{t+1} L_{t+1})^{4 / 3}+\frac{1}{2}\left(\frac{\eta_{t+1} L_{t+1}}{D}\right)^{2 / 3} u_{t+1} \\
&\stackrel{(f)}{\leq}& \frac{1}{4} D^2 \sigma_t^2+\frac{1}{3} \sigma_t u_{t+1},
\end{eqnarray*}
where inequality $(e)$ follows from the AM-GM inequality and $(f)$ follows from the choice of the step size sequence $\eta_t = \frac{D}{2L_tT^{\frac{3}{4}}},$ and the fact that $\sigma_t \geq \frac{1}{\sqrt{t}}$.  
To bound term (d) from Eqn.\ \eqref{term-d}, observe that,
$$
\begin{gathered}
D^2(1 - \frac{\eta_{t+1}}{\eta_t}) = D^2\big(1 - \frac{L_t}{L_{t+1}}\big). 
\end{gathered}
$$
Substituting the above bounds into Eqn.\ \eqref{term-d}, we have
\[
u_{t+1} \leq (1-\sigma_t)u_t + \frac{5}{4}D^2\sigma_t^2 + \frac{1}{3} \sigma_t u_{t+1} + D^2(1 - \frac{L_t}{L_{t+1}}) 
\]
This implies that
\begin{eqnarray*} 
 u_{t+1} \big(1-\frac{1}{3}\sigma_t\big) \leq (1-\sigma_t)u_t + \frac{5}{4}D^2\sigma_t^2 + D^2(1 - \frac{L_t}{L_{t+1}}),
\end{eqnarray*}
\emph{i.e.,}
\begin{eqnarray} \label{u-recursion}
	u_{t+1} \leq \big(1-\frac{2}{3}\sigma_t\big) u_t + \frac{15}{8}D^2 \sigma_t^2 + \frac{3}{2}D^2 \big(1-\frac{L_t}{L_{t+1}}\big),
\end{eqnarray}
where we have used the fact that $\sigma_t \leq 1$ and $\frac{1-\sigma_t}{1-\frac{1}{3}\sigma_t} \leq 1-\frac{2}{3}\sigma_t.$ Using the fact that the sequence $\{\sigma_t\}_{t\geq 1}$ is non-increasing and $u_{T+1}\geq 0,$ summing up the inequalities in \eqref{u-recursion} in the range $t=1$ to $t=T-1$, we conclude that 
\begin{eqnarray*}
\frac{2}{3}\sigma_T\sum_{t=1}^Tu_{t} &\leq& u_1+ \frac{15}{8}D^2\sum_{t=1}^{T-1}\sigma_t^2 + \frac{3}{2}D^2\sum_{t=1}^{T-1}\frac{L_{t+1}-L_t}{L_{t+1}} \\
&\leq & u_1 + \frac{15}{2}D^2 \sum_{t=1}^{T-1} \frac{1}{t} + \frac{3}{2}D^2 \int_{L_1}^{L_{T}} \frac{dx}{x} \\
&\leq & u_1 + \frac{15}{2}D^2(1+ \ln T) + \frac{3}{2}D^2 \log \frac{L_T}{L_1}.
\end{eqnarray*}
Since $\sigma_T \geq \frac{1}{\sqrt{T}},$ the above bound implies that 
\[
\sum_{t=1}^T u_t \leq \frac{3\sqrt{T}}{2}u_1 + \frac{45 D^2\sqrt{T}}{4}(1+\ln T) + \frac{9D^2 \sqrt{T}}{4} \log \frac{L_T}{L_1}.
\]
It can be easily verified that $u_1 =\eta_1 h_1 \leq D^2/2.$
Finally, we have
\[
\sum_{t=1}^T L_t ||x_t - x_{t+1}^\star|| \stackrel{(a)}{\leq} L_T \sum_{t=1}^T \sqrt{u_t} \stackrel{(b)}{\leq} L_T \sqrt{T} \sqrt{\sum_{t=1}^T u_t}  \leq 12 L_T D T^{3/4} \sqrt{\log(Te L_T/L_1)}),
\]
where inequality (a) follows upon combining the monotonicity of the Lipschitz constants, the bound from Eqn. \eqref{u-t-bd} and the definition of the sequence $\{u_t\}_{t\geq 1}$ and (b) follows from Cauchy-Schwarz inequality.

\subsection{Proof of Lemma \ref{term-d-lemma}} \label{term-d-lemma-pf}
Since $F_t$ is $\frac{2}{\eta_t}$-strongly-convex and $x^\star_{t+1} \in \arg\min_{x \in \mathcal{X}} F_t(x),$ we have
\begin{eqnarray*}
       &&\Vert x_{t}^\star-x_{t+1}^\star \Vert ^2/ \eta_{t} \leq F_{t}(x_{t}^\star) - F_{t}(x_{t+1}^\star) 
      \leq  \underbrace{F_{t-1}(x_{t}^\star)- F_{t-1}(x_{t+1}^\star)}_{\leq 0} +  \nabla_{t}^{\top} (x_{t}^\star - x_{t+1}^\star)   
      + \Vert x_{t}^\star-x_{1}^\star \Vert ^2(\frac{1}{\eta_{t}}-\frac{1}{\eta_{t-1}}), 
 \end{eqnarray*}     
      where we have used the fact that $\{\eta_t\}_{t \geq 1}$ is non-increasing. Hence, using Cauchy-Schwarz inequality, and using the fact that $\eta_t = \frac{D}{2L_tT^{\frac{3}{4}}}, t\geq 1$, we have 
      \begin{eqnarray*}
      &&\Vert x_{t}^\star-x_{t+1}^\star \Vert ^2\leq  \eta_t  \Vert \nabla_{t} \Vert  \Vert x_{t}^\star - x_{t+1}^\star \Vert + D^2(1-\frac{L_{t-1}}{L_{t}})\nonumber \\
      &\implies &\Vert x_{t}^\star-x_{t+1}^\star \Vert \leq 2\bigg(\eta_t  \Vert \nabla_{t} \Vert + D \sqrt{1-\frac{L_{t-1}}{L_{t}}}\bigg). \label{eq:absolut_dist_full}
    \end{eqnarray*}
    Define $\xi \equiv \max_{1\leq t \leq T} \frac{L_t}{L_{t-1}}.$ We can bound the summation as follows
    \begin{eqnarray*}
    	\sum_{t=1}^T L_t \sqrt{1- \frac{L_{t-1}}{L_t}} = \sum_{t=1}^T \sqrt{L_t(L_t-L_{t-1})} \leq \sqrt{L_T} \sum_{t=1}^T \sqrt{L_t-L_{t-1}} \leq L_T \sqrt{T},
    \end{eqnarray*}
    where the last step follows from an application of the Cauchy-Schwarz inequality.
    Hence, 
    \begin{eqnarray}
        \sum_{t=1}^T L_t||x_{t}^\star-x_{t
        +1}^\star|| \leq D L_T T^{\nicefrac{1}{4}} + 2D L_T \sqrt{T} \leq 3DL_{T}\sqrt{T}.
    \end{eqnarray}

\iffalse
\[
\text{Regret}_T \leq 3 ~\textrm{term (a)} + \textrm{Regret}_T(\textrm{FTRL})
\]
\[
\text{Regret}_T \leq 3 ~\textrm{term (a)} +\frac{D^2}{\eta_{T-1}} + \frac{1}{2} \sum\limits_{t=1}^{T}\eta_{t-1} L_t^2
\]
\[
\text{Regret}_T \leq 3 L_T D T^{3/4} \sqrt{\log(TL_T)} +\frac{D^2}{\eta_{T-1}} + \frac{1}{2} \sum\limits_{t=1}^{T}\eta_{t-1} L_t^2
\]
\[
\text{Regret}_T \leq 5 L_T D T^{3/4} \sqrt{\log(TL_T)} 
\]
\fi

\subsection{Proof of Lemma \ref{ratio-bd}} \label{ratio-bd-proof}
Recall that the sequence $\{Q(t)\}_{t \geq 1}$ is monotone non-deceasing with $Q(0) = GD\log T$. Furthermore, since $g_t(x^\star)\leq 0,$ for any $t,$ from the Lipschitz continuity of the constraint function $g_t,$ we have $g_t(x_t) \leq g_t(x^\star) + G||x_t -x^\star|| \leq GD.$ This shows that $Q(t)-Q(t-1) \leq GD.$
From the expression of the Lipschitz constant of the surrogate cost functions, given by Eqn. \eqref{lipschitz-const}, we have:
\[
L_t = G\big(V + p Q^{p-1}(t)\big), ~~ \textrm{and} ~~ L_{t-1} = G\big(V + p Q^{p-1}(t-1)\big).
\]
Hence, the ratio of the successive Lipschitz constants can be bounded as:
\begin{align*}
    \frac{L_t}{L_{t-1}} &=  \frac{V + p Q^{p-1}(t)}{V + p Q^{p-1}(t-1)} \\
    &= 1 + \frac{p (Q^{p-1}(t)-Q^{p-1}(t-1))}{V + p Q^{p-1}(t-1)} \\
    &\stackrel{(a)}{\leq} 1 + GDp(p-1)\frac{Q^{p-2}(t)}{V + p Q^{p-1}(t-1)} \\
    &\stackrel{(b)}{\leq} 1 +GDp \frac{\big(Q(t-1)+GD\big)^{p-2}}{ Q^{p-1}(t-1)} \\
    &= 1 + \frac{GDp}{Q(t-1)}\bigg(1+ \frac{GD}{Q(t-1)}\bigg)^{p-2}\\
    &\stackrel{(c)}{\leq} 1 +  \bigg(1 + \frac{1}{p}\bigg)^{p-2} \\
    &\stackrel{(d)}{\leq} 1 +  e^{1-\frac{2}{p}}\\
    &\leq 1+e
\end{align*}
where (a) follows from the convexity of the mapping $x\mapsto x^{p-1}$, which yields 
\begin{eqnarray*}
	Q^{p-1}(t-1) \geq Q^{p-1}(t) +  (p-1)Q^{p-2}(t) (Q(t-1)-Q(t)),
\end{eqnarray*}
\emph{i.e.,}
\begin{eqnarray*}
	Q^{p-1}(t)- Q^{p-1}(t-1) \leq (p-1)Q^{p-2}(t)(Q(t)-Q(t-1)) \leq (p-1)GD Q^{p-2}(t),
\end{eqnarray*}
where we have used the fact that $Q(t)-Q(t-1) \leq GD.$  
Inequality (b) uses the previous bound and the fact that 
$V$ is non-negative. Inequality (c) follows from  the fact that $Q(t-1)\geq Q(0) = GD\log T=GDp.$ Finally, inequality (d) follows from the fact that $e^t \geq 1+t, \forall t \in \mathbb{R}.$ This shows that for any $T \ge 1,$ we have
\[ \max_{1\leq t \leq T} \frac{L_{t}}{L_{t-1}} \leq 1+e. \]

\subsection{Proof of Lemma \ref{sqrt-ratio}} \label{sqrt-ratio-proof}

Recall that the sequence $\{Q(t)\}_{t \geq 1}$ is monotone non-deceasing with $Q(0) = GD\log T$. Furthermore, the Lipschitz constants of the surrogate costs functions depend on the CCV as
\[
L_T = G\big(V + p Q^{p-1}(T)\big), ~~ \textrm{and} ~~ L_{1} = G\big(V + p Q^{p-1}(1)\big),
\]
where $p=\log T.$
Hence,
\begin{eqnarray*}
    \frac{L_T}{L_{1}} =  \frac{V + p Q^{p-1}(T)}{V + p Q^{p-1}(1)} 
    \stackrel{(a)}{\leq} \frac{Q^{p-1}(T)}{Q^{p-1}(1)} 
    \stackrel{(b)}{\leq} \frac{(GDT + GD\log T)^{p-1}}{(GD\log T)^{p-1}} 
    \stackrel{(c)}{\leq} T^{p}. 
\end{eqnarray*}
where inequality (a) follows from the fact that $Q(T) \geq Q(1), V>0$ and the map $x \mapsto \frac{x+\alpha}{x+\beta}$ is non-increasing for $\alpha\geq \beta$, (b) uses the fact that $Q(t)-Q(t-1) \leq GD, t \geq 1,$ and $Q(0)=GD \log T,$ and finally, (c) holds for $T\geq 2$.
Hence, we have
\begin{align*}
    \log \frac{L_T}{L_{1}} \leq  p(\log T)  \implies 
     \sqrt{\log \frac{L_T}{L_{1}}} \leq \log T,
\end{align*}

\section{The Bandit Feedback Setting}
\label{bandit_feedback}
\subsection{Adaptive Blocked Bandit Conditional Gradient algorithm}
\label{bandit-background}
\citet{hazan2022introduction} presents a simple reduction from the bandit feedback setting to the full information setting for first-order algorithms. The reduction guarantees the same regret bounds as the original first-order algorithm up to
the magnitude of the estimated gradients. In other words, as long as we can reasonably bound the variance of the gradient estimator, we should expect the same bounds as the analogous full-information algorithm. It turns out that while the reduction applies to the OCG algorithm by virtue of it being a first-order algorithm, the resulting bounds can be further improved by modifying the algorithm to control the variance of the gradient estimators better.

To achieve that end, we employ a standard tool known as the \textit{blocking} technique \cite{hazan2020faster,garber2019improvedregretboundsprojectionfree, garber2024projectionfree, 1013135}. This is the main departure from section \ref{ocg-section}, apart from the obvious transition from full information to a bandit setting. The blocking technique is a general technique where the horizon is partitioned into non-overlapping blocks of length $K$. Instead of playing a new action at every timestep, the policy plays the same action throughout each block $m$ while switching to a new action at the start of block $m+1$. 

\begin{remark}
\label{remark:block-nonblock}
For the sake of analysis, it would be useful to remember that this technique is essentially the grouping of several iterations of the sequential game into one. This observation would enable a reduction from a blocking algorithm to a non-blocking algorithm.
\end{remark}

The purpose of the blocking technique is variance-reduction and it is closely related with another technique known in literature as the averaging technique \cite{mokhtari2017conditionalgradientmethodstochastic, chen2018projectionfreeonlineoptimizationstochastic}. Perhaps the latter better alludes to the intuition behind this technique. As the gradients corresponding to the functions in the block are accumulated, the corresponding variance of the sum of their estimators only scales linearly with $K$ instead of $K^2$. This argument is formalized in Lemma \ref{lemma:expectation_gradient}.

We first introduce the \textit{smoothed} version of the cost function whose value at a point $x$ is the same as the expected value of the original cost function over all points in the $\delta$-neighbourhood of the point $x$. These play a crucial role in the analysis. The reason this becomes important is because we do not play a constant action throughout a block $m$ but a $\delta$-perturbation of it. This is essential as it is a well-known fact that a bandit algorithm has to be randomised. So the value of the smoothed function at the point $x$ is a better representative of the expected cost than the value of the actual cost function would be. Note that these functions don't appear in the algorithm and are only defined for the sake of analysis. Formally,
 
 \begin{equation}
 \label{smoothed-function}
    \hat{f}_{\delta}(\mathbf{x}) = \mathbb{E}_{\mathbf{u} \sim \ball^n} \left[ f(\mathbf{x} + \delta \mathbf{u}) \right]
\end{equation}
 \begin{equation}
 \label{smoothed-function-gradient}
    \nabla\hat{f}_{\delta}(\mathbf{x}) = \mathbb{E}_{\mathbf{u} \sim \ball^n} \left[ \frac{n}{\delta}f(\mathbf{x} + \delta \mathbf{u})\mathbf{u} \right]
\end{equation}
Also, as in \cite{flaxman2004online} we make the standard assumption that the feasible set $\mathcal{X}$ is full dimensional, contains the origin, and that there exists scalars $r,R>0$ such that $r\ball^n\subseteq\mathcal{X}\subseteq{}R\ball^n$, where $\ball^n$ denotes the unit Euclidean ball centered at the origin in $\reals^n$.

We denote by $\mathcal{S}^n$ the unit sphere in $\mathbb{R}^n$, and we write $u \sim S^n$ and $u \sim \ball^n$ to denote a random vector $u$ sampled uniformly from $\mathcal{S}^n$ and $\ball^n$, respectively. We denote by $\Vert \mathbf{x} \Vert$ the $\ell_2$ norm of the vector $\mathbf{x}$. %We denote $\mathcal{K}_{\delta} \subseteq \mathcal{K}$ when $\forall \mathbf{x} \in \mathcal{K}_{\delta}, \mathbf{u} \sim B^n$ imply that $\mathbf{x} + \delta \mathbf{u} \in \mathcal{K}$.

For a compact and convex set $\mathcal{X}\subset\reals^n$, which satisfies the above assumptions (i.e., $r\ball^n\subseteq\mathcal{X}\subseteq{}R\ball^n$), and a scalar $0 < \delta \leq r$, we define the set $\mathcal{X}_{\delta} := (1-\delta/r)\mathcal{X} = \{(1-\delta/r)x~|~x\in\mathcal{X}\}$. In particular, it holds that $\mathcal{X}_{\delta}\subseteq\mathcal{X}$ and for all $x\in\mathcal{X}_{\delta}$, $x+\delta\ball^n\subseteq\mathcal{X}$ (see \cite{HazanBook}). This is important as we want to make sure that a $\delta$-perturbation around $x_m$, that is, the constant expected action in a block, does not land us outside $\mathcal{X}$. This means that $x_m$ should belong to $\mathcal{X}_{\delta}$ and that the optimization in Algorithm \ref{alg:Cg} is over $\mathcal{X}_{\delta}$ as opposed to $\mathcal{X}$.

Finally, note that while the conditional gradient method is run over the shrunk set $\mathcal{X}_{\delta}$, solving the linear optimization problem over $\mathcal{X}_{\delta}$ is identical, up to scaling, to solving it over the original set $\mathcal{X}$. 

\begin{remark}In line with \cite{HazanBook} much of the analysis of a BCO algorithm can be reduced to that of an OCO algorithm by simply replacing the cost and constraint functions by their $\delta-$smoothed versions.\end{remark}
\subsection{Conditional Gradient with Stopping Condition}
\label{cond-grad-sec}
\begin{algorithm}
\caption{Conditional Gradient with Stopping Condition}\label{alg:Cg}
\begin{algorithmic}[1]
\STATE \textbf{Inputs:} Convex decision set $\mathcal{X}_{\delta},$ initial vector $x_{\text{in}},$ objective function $F_m(x).$
\STATE \textbf{Parameters:} error tolerance $\epsilon$.
\STATE Initialize $z_1 \leftarrow{} x_{in}$ and $\tau \leftarrow{} 0$
\REPEAT
\STATE $\tau \leftarrow{} \tau +1$
\STATE $v_{\tau} \in \argmin\limits_{x \in \mathcal{X}_{\delta}} \{ \nabla F_m(z_\tau)^{\top} x \} $
\STATE $ \sigma_{\tau} = \argmin\limits_{\sigma \in [0, 1]}  \{ F_m(z_\tau + \sigma (v_\tau - z_\tau)) \} $
\STATE $z_{\tau+1} = z_\tau + \sigma_{\tau} (v_\tau - z_\tau)$
\UNTIL{$\nabla F_m(z_{\tau})^{\top}(z_{\tau}-v_{\tau}) < \frac{\epsilon}{\eta_m}$} \label{line 9}
\STATE $x_{\text{out}} \xleftarrow{} z_{\tau}$
\end{algorithmic}
\end{algorithm}

The above Algorithm \ref{alg:Cg} serves as a subroutine to Algorithm \ref{our_alg_bandit} which is our adaptive bandit algorithm. 

The algorithm tries to find the approximate minimizer of the function $F$ by using only LOO calls. The process repeats until the Frank-Wolfe gap, defined as $g_\tau = \nabla F_m(z_{\tau})^\top(z_{\tau} - v_{\tau})$, falls below a specified tolerance threshold, $\epsilon$. The gap $g_\tau$ provides a computable upper bound on the suboptimality of the current iterate, i.e., $F_m(z_\tau) - F_m(x^*) \leq g_\tau$, where $x^*$ is the true optimum.

\subsection{Proof of Theorem \ref{thm:main-bandit}}
\label{thm-proof}
\iffalse
\begin{theorem}  \label{thm:main-restated}
    %Let $L_m$ the number of calls in block $m$ to the linear optimization oracle. 
    Setting $\eta_m = \frac{2 R}{G_m} T^{-\frac{3}{4}}$, $\delta = \sqrt{n} \frac{r}{R}T^{-\frac{1}{4}} $, $K = T^{\frac{1}{2}}$ in Algorithm \ref{alg:VRBCO}, guarantees that the expected regret is upper-bounded by
    {\small\begin{align}
        \mathbb{E}[\mathcal{R}_{T}] \leq & 4\sqrt{n}L_TT^{\frac{3}{4}}+3T\sqrt{\epsilon} + D G_{T/K}\big(1 + \max_{1\leq m\leq T/K} \frac{G_m}{G_{m-1}}\big) T^{3/4} , \nonumber
    \end{align}}
    and that the expected overall number of calls to the linear optimization oracle is upper-bounded by
    \begin{align}
        \mathbb{E} \left[ \sum_{m=1}^{\frac{T}{K}} N_m \right] \leq & ~ \frac{16R^4}{\epsilon^2} \left ( \frac{\sqrt{\epsilon}}{R} T^{1/4} + 1 + \sqrt{\log \frac{G_{T/K}}{G_1}}\right) . \nonumber
    \end{align}
\end{theorem}
\fi

Define $x^* \in \argmin\limits_{x \in  \mathcal{X}} \sum_{t=1}^{T} f_t(x)$, $\tilde{x}^* =  (1- \delta/r)x^*$, $\forall t : m(t) := \lceil\frac{t}{K}\rceil$. Recall that throughout any block $m$,  Algorithm \ref{alg:VRBCO} predicts according to  $x_{m}$. 

It is clear from Algorithm \ref{alg:VRBCO} that there is a lot of randomness involved which if separated would considerably ease our analysis. Note that what we are about to do next can be considered an implicit reduction from the bandit to full information setting.

To that end we define a proxy regret - 
\begin{equation}
\label{proxy-regret}
\textrm{Regret}_T\textrm{(blocked-OCG)}   = \sum\limits_{t=1}^{T} \big(\E[\hat{f}_{t,\delta}(x_{m(t)})] - \hat{f}_{t,\delta}(\tilde{x}^\star)\big)
\end{equation}
The following lemma relates the true expected regret with the regret defined in Eqn. \eqref{proxy-regret}.
\begin{lemma}
\label{lemma:bound_proxy_regret}
Under Algorithm \ref{alg:VRBCO}, the true expected regret can be upper bounded as 
\begin{eqnarray}
\E[\textrm{Regret}_T(\textrm{Algorithm \ref{alg:VRBCO}})] \leq& 3L_T \delta T + \frac{\delta R L_T T}{r} + \textrm{Regret}_T\textrm{(blocked-OCG)}\nonumber
\end{eqnarray}
\end{lemma}
The proof of Lemma \ref{lemma:bound_proxy_regret} is given in Appendix \ref{bound_proxy_regret_pf}.
\paragraph{Bounding $\textrm{Regret}_T\textrm{(blocked-OCG)}$ :}
\begin{lemma}
\label{bound-term-c}
    Under Algorithm \ref{alg:VRBCO}, $\textrm{Regret}_T\textrm{(blocked-OCG)}$ can be upper bounded as
	\begin{eqnarray*}
		\frac{4R^2}{\eta_{T/K}} + \frac{1}{4} \sum\limits_{m=1}^{T/K}\eta_{m-1} \E[||\nabla_{m,\delta}||^2] + 3L_T T\sqrt{\epsilon}
	\end{eqnarray*}
\end{lemma}
The proof of this lemma is given in \ref{bound-term-g-proof}.

We next set $\eta_m = \frac{2R}{K(\frac{nM_{mK}}{\delta} + L_{mK}\sqrt{K})}$ and expanding $\E[||\nabla_{m,\delta}||^2]$ using Lemma \ref{lemma:expectation_gradient}. We also set $\delta = \sqrt{n}T^{-1/4}$ and $K = \sqrt{T}$.

Overall, we get,
\begin{align}
\label{regret-bound-bco}
\E[\textrm{Regret}_T(\textrm{Algorithm \ref{alg:VRBCO}})] \leq & \sqrt{n}L_T (3 + \frac{ R}{r}) T^{3/4} + 3L_T T\sqrt{\epsilon}  \nonumber \\ + &2R(\sqrt{n}M_{T} + L_T)\big(1 + \max_{1\leq m\leq T/K} \frac{\sqrt{n}M_{mK} + L_{mK}}{\sqrt{n}M_{(m-1)K} + L_{(m-1)K}}\big) T^{3/4} 
\end{align}
 
\paragraph{Bounding the total number of calls to the oracle:}

We begin by mentioning the following lemma-
\begin{lemma}
    \label{lemma:total-calls-LOO}
    Under Algorithm \ref{alg:VRBCO}, the expected overall number of calls to the linear optimization oracle is upper-bounded by
    \begin{align}
    \label{num-calls}
        \mathbb{E} \left[ \sum_{m=1}^{\frac{T}{K}} N_m \right] \leq & ~ \frac{16R^4}{\epsilon^2} \left ( \frac{\sqrt{\epsilon}}{R} T^{1/4} + 1 + \sqrt{\log \frac{\sqrt{n}M_{T} + L_T}{\sqrt{n}M_1+L_1} }\right) . 
    \end{align}
\end{lemma}
The proof of the lemma is given in \ref{total-calls-LOO-proof}.

We can further bound the RHS of 
 Eqn. \eqref{num-calls} by $T$ by choosing $\epsilon$ appropriately. 

\subsection{Proof of Theorem \ref{perf_thm_bandit}}
\label{bandit-reg-vio}

We have to bound $\E\textrm{Regret}'_T$. 

\begin{align*}
    \E\textrm{Regret}'_T \leq \E[  (3+\frac{R}{r})\sqrt{n}L_TT^{\frac{3}{4}}+3L_T T\sqrt{\epsilon} \nonumber + 2R G_T\big(1 + \max_{1\leq m\leq T/K} \frac{G_{mK}}{G_{(m-1)K}}\big) T^{3/4}]
\end{align*}

and also, we have to bound 
\begin{align*}
    \mathbb{E} \left[ \sum_{m=1}^{\frac{T}{K}} N_m \right] \leq \frac{16R^4}{\epsilon^2} \E\left ( \frac{\sqrt{\epsilon}}{R} T^{1/4} + 1 + \sqrt{\log \frac{G_{T}}{G_1}}\right)
\end{align*}
where $G_t = \sqrt{n}M_t + L_t.$

Now, note that $M_t = (V+pQ^{p-1}(T))F$ where $F$ is the sup-norm of cost and constraint functions. $L_t = (V+pQ^{p-1}(T))G$ where $G$ is the Lipchitz constant of cost and constraint functions. So, we have $G_t = (V+pQ^{p-1}(T))(\sqrt{n}F+G)$

We must first find the appropriate value of $\epsilon$ for which the expected total number of calls to the LOO is $T.$ First, note that $\log \frac{G_{T/K}}{G_1} \leq \log \frac{V + pQ^{p-1}(T)}{V+ pQ^{p-1}(1)} \leq  \log \frac{Q^{p-1}(T)}{Q^{p-1}(1)} \leq \log Q^{p-1}(T) \leq \log T^{p-1} \leq (\log T)^2.$ Therefore, the RHS of the above inequality is bounded by,

\begin{align*}
  \frac{16R^4}{\epsilon^2}\left( \frac{\sqrt{\epsilon}}{R} T^{1/4} + 1 + \log T\right)
\end{align*}
we can bound this by $T$, by taking $\epsilon = 16R^2T^{-1/2}\sqrt{\log T}.$

Now, let us look at the $\max_{1\leq m\leq T/K} \frac{G_m}{G_{m-1}}$ term in the inequality bounding $\E\textrm{Regret}'_T$. $\frac{G_m}{G_{m-1}} \leq \frac{V + pQ^{p-1}(mK)}{V + pQ^{p-1}((m-1)K)} \leq \frac{Q^{p-1}(mK)}{Q^{p-1}((m-1)K)} \leq (1 + \frac{KGD}{Q(0)})^p \leq e.$ The last inequality follows because $Q(0) = KGD\log T = \sqrt{T}GD \log T.$ Therefore, we get,

\begin{align*}
    \E\textrm{Regret}'_T \leq \E[ (3+\frac{R}{r})\sqrt{n}L_TT^{\frac{3}{4}}+12L_T RT^{3/4}(\log T)^{1/4} \nonumber + 2R G_T\big(1 + e\big) T^{3/4}] 
\end{align*}
We overall get,

\begin{align*}
    \E\textrm{Regret}'_T \leq  (30 + \frac{R}{r})\sqrt{n} (V+ \E\Phi'(Q(T)))T^{3/4} (\log T)^{1/4}
\end{align*}
Therefore, from the regret decomposition inequality \ref{bandit-reg-decomp} and setting $c = (30 + \frac{R}{r}) \sqrt{n}$,
\begin{align*}
    \mathbb{E}\Phi(Q(T)) - \Phi(Q(0)) + V\mathbb{E}\textrm{Regret}_T \leq c(V+ \E\Phi'(Q(T)))T^{3/4} 
\end{align*}
Substituting, $\Phi(x) = x^p$, we get,
\begin{align*}
    \mathbb{E}Q^p(T) - Q^p(0) + V\mathbb{E}\textrm{Regret}_T \leq cVT^{3/4} + cp\E Q^{p-1}(T)T^{3/4} 
\end{align*}
Bounding the Regret,
\begin{align*}
  \mathbb{E}\textrm{Regret}_T \leq cT^{3/4} + \frac{1}{V}\E[ \underbrace{cpQ^{p-1}(T)T^{3/4} - Q^p(T)}_{\leq (cpT^{3/4})^p}] + \frac{Q^p(0)}{V}
\end{align*}

If we choose $V= (cpT^{3/4})^p$ then we get,

\begin{align*}
\mathbb{E}\textrm{Regret}_T \leq cT^{3/4} + 2
\end{align*}
To bound the CCV, consider that $\textrm{Regret}_T \geq -T$,
\begin{align*}
  \mathbb{E}Q^p(T)  \leq 2cVT + cm\E Q^{p-1}(T)T^{3/4} + Q^p(0) 
\end{align*}

Either of the following two would hold in that case, 
\begin{align*}
  \mathbb{E}Q^p(T)  \leq 4cVT + 2Q^p(0) \stackrel{(a)}{\implies} \mathbb{E}Q(T)^p \leq 4cVT + 2Q^p(0) \implies \mathbb{E}Q(T) &\leq (4cVT)^{1/p} + 2Q(0) \nonumber \\ &\leq cpT^{3/4} (4cT)^{1/p} + 2GD\sqrt{T} \log T
\end{align*}

\begin{align*}
  \mathbb{E}Q^p(T)  \leq 2cp\E Q^{p-1} T^{3/4} \stackrel{(b)}{\implies} [\mathbb{E}Q^{p-1}(T)]^{\frac{p}{p-1}}  \leq 2cp\E Q^{p-1} T^{3/4} &\implies [\mathbb{E}Q^{p-1}(T)]^{\frac{1}{p-1}} \leq 2cpT^{3/4} \nonumber \\ &\stackrel{(c)}{\implies} \mathbb{E} Q(T) \leq 2cpT^{3/4}
\end{align*}

where (a) follows from the convexity of the map $x \mapsto x^{p}$ and using Jensen's inequality and $(b)$ follows from applying Jensen's inequality on the LHS to the convex map $x \mapsto x^\frac{p}{p-1}$, resulting in 
\[\mathbb{E}(Q^p(T)) = \mathbb{E}\big[\big(Q^{p-1}(T)\big)^{\frac{p}{p-1}}\big] \geq \mathbb{E}\big[Q^{p-1}(T)\big]^{\frac{p}{p-1}},\]
(c) follows from the convexity of the map $x \mapsto x^{p-1}$ and using Jensen's inequality.

Thus, both expected regret and violation are $\tilde{O}(T^{3/4}).$
\subsection{Proof of Lemma \ref{lemma:bound_proxy_regret}}
\label{bound_proxy_regret_pf}
\begin{eqnarray} \label{expected_regret_bd}
&&\E[\textrm{Regret}_T(\textrm{Algorithm \ref{alg:VRBCO}})] \nonumber\\ &=&\sum\limits_{t=1}^{T} \E[\big(\hat{f}_t(y_t)] - \hat{f}_t(x^\star)\big)  \nonumber\\
    &=& \underbrace{\sum\limits_{t=1}^{T} \big(\E[\hat{f}_t(y_t)] - \E[\hat{f}_t(x_{m(t)})]\big)}_{(A)}
    +\underbrace{\sum\limits_{t=1}^{T} \big(\hat{f}_t(\tilde{x}^\star) - \hat{f}_t(x^\star)\big)}_{(B)}
    +\underbrace{\sum\limits_{t=1}^{T} \big(\E[\hat{f}_t(x_{m(t)})] - \hat{f}_t(\tilde{x}^\star)\big)}_{(C)},
\end{eqnarray}
Bounding term (A):
\begin{eqnarray}
&&\sum\limits_{t=1}^{T} \big(\E[\hat{f}_t(y_t)] - \E[\hat{f}_t(x_{m(t)})]\big)
= \sum\limits_{t=1}^{T} \big(\E[\hat{f}_t(x_{m(t)} + \delta u_t) - \hat{f}_t(x_{m(t)})]\big) \leq \sum\limits_{t=1}^{T} L_t  \E[\Vert\delta u_t\Vert]  \nonumber \\ &\leq& L_T \delta T
\end{eqnarray}
Bounding term (B):
\begin{eqnarray}
    &&\sum\limits_{t=1}^{T} \big(\hat{f}_t(\tilde{x}^\star) - \hat{f}_t(x^\star)\big) 
    = \sum\limits_{t=1}^{T} \big(\hat{f}_t((1-\frac{\delta}{r})x^\star) - \hat{f}_t(x^\star)\big)  \leq \sum\limits_{t=1}^{T} L_t \Vert(1-\frac{\delta}{r})x^\star - x^\star\Vert \nonumber \\ &\leq& \frac{\delta R L_T T}{r}
\end{eqnarray}
Bounding term (C):
\begin{eqnarray}
    &&\sum\limits_{t=1}^{T} \big(\E[\hat{f}_t(x_{m(t)})] - \hat{f}_t(\tilde{x}^\star)\big) \nonumber \\
    &\leq& \sum\limits_{t=1}^{T} \big(\E[\hat{f}_{t,\delta}(x_{m(t)})] - \hat{f}_{t,\delta}(\tilde{x}^\star)\big) + \sum\limits_{t=1}^{T} \E[\underbrace{\big(\hat{f}_{t}(x_{m(t)}) - \hat{f}_{t,\delta}(x_{m(t)})\big)}_{\leq L_T\delta}] + \sum\limits_{t=1}^{T} \E[\underbrace{\big(\hat{f}_{t,\delta}(\tilde{x}^\star) - \hat{f}_{t}(\tilde{x}^\star)\big)}_{\leq L_T\delta}] \nonumber \\
    &\leq& \sum\limits_{t=1}^{T} \big(\E[\hat{f}_{t,\delta}(x_{m(t)})] - \hat{f}_{t,\delta}(\tilde{x}^\star)\big) + 2\delta L_TT = \textrm{Regret}_T\textrm{(blocked-OCG)} + 2\delta L_TT
\end{eqnarray}
\subsection{Proof of Lemma \ref{bound-term-c}} \label{bound-term-g-proof}
Define $x_m^{\star} = \argmin\limits_{x} F_{m-1}(x)$.

Where $F_m$ is defined in Eqn. \eqref{surr_fn_bandit}.
Using similar arguments as in the proof of Theorem \ref{ocg-reg-thm}, we get the following - 
\begin{eqnarray}
    \sum\limits_{t=1}^{T} \big(\E[\hat{f}_{t,\delta}(x_{m(t)})] - \hat{f}_{t,\delta}(\tilde{x}^\star)\big) \leq 3\underbrace{\sum\limits_{t=1}^{T} L_t\E[\Vert x_{m(t)} - x_{m(t)}^{\star} \Vert]}_{(A)} + \underbrace{\sum\limits_{t=1}^{T} \big(\E[\tilde{f}_{t,\delta}(x_{m(t)}^{\star})] - \tilde{f}_{t,\delta}(\tilde{x}^\star)\big)}_{(B)} 
\end{eqnarray}

It is easy to observe the similarity between the above equation and the Eqns. \eqref{regret_bd} and \eqref{intermediate_bd}. The only difference is the use of the expectation operator which is an artefact of the randomised nature of the bandit algorithm. $\tilde{f}_{t,\delta}$ is the smoothed analogue of $\tilde{f}_{t}$ which is as defined in Eqn. \eqref{f_tilde}.

In this case bounding term (A) wouldn't be as tedious as it was in the case of Algorithm \ref{ocg_alg}. This is because Algorithm \ref{alg:Cg} (line 9) ensures that it remains bounded by $\sqrt{\epsilon}$. This is demonstrated in the following inequalities.
    \begin{align}
       \frac{\epsilon}{\eta_{m-1}} \geq g_{m-1}^{\top} (x_{m} - v_{L_m}) \geq g_{m-1}^{\top} (x_{m} - x^*_{m}) 
        \geq & F_{m-1}(x_{m}) - F_{m-1}(x^*_{m}) \geq \frac{\Vert x_{m} - x^*_{m}\Vert^2}{\eta_{m-1}} \label{eq:aux1_cg_lemma_prev}
    \end{align}
Algorithm \ref{alg:Cg} ensures the first inequality by virtue of its loop condition. Where $g_{m-1} = \nabla F_{m-1}(x_{m})$ and $v_{L_m}$ denotes the value of $v_{\tau}$ in the last iteration of the loop in Algorithm \ref{alg:Cg}. The second inequality follows from the fact that $v_{L_m}$ is the minimiser of $g_{m-1}^{\top} x$. The third inequality follows from the convexity of $F_m$. The last inequality follows from the $\frac{2}{\eta_m}$-strong convexity of $F_m$.
Now we bound term (B) -
\begin{eqnarray}
\label{blocked-ftrl}
    (B)\equiv \sum_{m=1}^{T/K} \big(\E[\tilde{f}_{(m-1)K+1:mK,\delta}(x_{m}^{\star})] - \tilde{f}_{(m-1)K+1:mK,\delta}(\tilde{x}^\star)\big)
\end{eqnarray}
The $:$ notation is used to denote accumulation and $f_{t_1:t_2}$ denotes the sum of losses from $t_1$ to $t_2$ (both inclusive). Eqn. \eqref{blocked-ftrl} makes it clear that term (B) is essentially the regret of the blocked FTRL algorithm where the same notion of blocking is used. This is essentially a reduction from a blocking algorithm to a non-blocking one as mentioned in Remark \ref{remark:block-nonblock}. We appeal to Theorem \ref{ftrl_bound}.
\begin{eqnarray}
\label{blocked-ftrl-regret}
    (B) \leq \frac{D^2}{\eta_{T/K}} + \frac{1}{4} \sum\limits_{m=1}^{T/K}\eta_{m-1} \E[||\nabla_m||^2]
\end{eqnarray}
In this case, $\nabla_m$ is used the denote the accumulated gradients of the block. We further appeal to Lemma \ref{lemma:expectation_gradient} and substitute the value of $\eta_m$, $\delta$ and $K$ and also noting that both $\{M_t\}_{1\leq t \leq T}$ and $\{L_t\}_{1\leq t \leq T}$ are non-decreasing sequences.
\subsection{Proof of Lemma \ref{lemma:total-calls-LOO}}
\label{total-calls-LOO-proof}
\begin{remark} Note that the function $\eta_m F_m$ is directly applicable to Lemma \ref{lemma:cg_epsilon_error_L_iterations} and scaling $F_m$ by $\eta_m$ does not affect the optimization in Algorithm \ref{alg:Cg} either. Henceforth, by $F_m$ we would denote the scaled version of the previous $F_m$. Thus it will be $2$-smooth and $2$-strongly convex. \end{remark}

Let $z_{m,\tau}$ be the iterate of Algorithm \ref{alg:Cg} after completing $\tau-1$ iterations of the do-while loop, when invoked on iteration (block) $m$ of Algorithm \ref{alg:VRBCO}. Also, for all $m,\tau$, define $h_{m,\tau} := F_{m}(z_{m,\tau}) -  F_{m}(x_{m+1}^*)$. Recall that for any iteration $m$ of Algorithm \ref{alg:VRBCO}, we have $z_{m,1} = x_{m}$.

Using the triangle inequality and the fact $F_{m-1}(x_{m+1}^*) \geq F_{m-1}(x_{m}^*)$, we have
\begin{align}
    \mathbb{E}[h_{m,1}] & =  \mathbb{E}[F_{m}(z_{m,1}) -  F_{m}(x_{m+1}^*)] 
    \leq  \mathbb{E}  [  F_{m-1}(x_{m}) -  F_{m-1}(x_{m}^*) + \eta_m   \Vert \nabla_{m} \Vert ~  \Vert  x_{m} - x_{m+1}^* \Vert ]. \nonumber
\end{align}
Since $h_{m-1,N_{m-1}} = F_{m-1}(x_{m}) -  F_{m-1}(x_{m}^*) \leq \epsilon$, using the triangle inequality, we have
\begin{align}
    \mathbb{E}[h_{m,1}]  
    \leq  \epsilon +  \eta_m \mathbb{E}[  \Vert \nabla_{m} \Vert \Vert x_{m} - x_{m}^* + x_{m}^* - x_{m+1}^* \Vert ]
    \leq \epsilon + \eta_m \mathbb{E}[    \Vert \nabla_{m} \Vert  \left(  \Vert x_{m} - x_{m}^* \Vert  +  \Vert x_{m}^* - x_{m+1}^* \Vert  \right) ]. \nonumber
\end{align}
Since $F_{m}(x)$ is $2$-strongly convex and $h_{m,L_{m}} \leq \epsilon$, we have that $ \Vert x_{m} - x_{m}^* \Vert  \leq \sqrt{\epsilon}$. Also, from Eq. \eqref{eq:absolut_dist}, we have $ \Vert x_{m}^* - x_{m+1}^* \Vert  \leq \eta_m  \Vert \nabla_{m} \Vert + D \sqrt{1-\frac{G_{m-1}}{G_{m}}}$. Thus, we have
\begin{align}
    \mathbb{E}[h_{m,1}] \leq  \epsilon  +  \eta_m \sqrt{\epsilon}~ \mathbb{E}[ \Vert \nabla_{m} \Vert ]  + \eta_m^2 \mathbb{E}[ \Vert \nabla_{m} \Vert ^2]  + \eta_m \mathbb{E}[ \Vert \nabla_{m} \Vert ] D \sqrt{1-\frac{G_{m-1}}{G_{m}}}. \nonumber
\end{align}
Using Lemma \ref{lemma:expectation_gradient} and the fact that for all $a,b \in \reals^+$ it holds that that $\sqrt{a+b} \leq \sqrt{a} + \sqrt{b}$, we have
\begin{align}
    \mathbb{E}[h_{m+1,1}] \leq  \epsilon  + \eta_m  \sqrt{\epsilon} ~ \left( \sqrt{K} \left(\frac{n M_{mK}}{\delta}\right) + K L_{mK} \right) + \eta_m^2 \left(K \left(\frac{n M_{mK}}{\delta}\right)^2 + K^2 L_{mK}^2 \right) \nonumber \\+ \eta_m D \sqrt{1-\frac{G_{m-1}}{G_{m}}} \left( \sqrt{K} \left(\frac{n M_{mK}}{\delta}\right) + K L_{mK} \right). \label{eq:upper_bound_h_1_cg_lemma}
\end{align}
Using Lemma \ref{lemma:cg_epsilon_error_L_iterations} with $\tilde{\epsilon}$ as the RHS of \eqref{eq:upper_bound_h_1_cg_lemma} for $m = 1, \dots, \frac{T}{K}$, we have that on each iteration (block) $m$, the number of calls to the linear optimization oracle is $N_m \leq \max \bigg{\{} \frac{16R^2}{\epsilon^2} (h_{m,1} - \epsilon) ,~ \frac{2}{\epsilon} (h_{m,1} - \epsilon) \bigg{\}}$. Plugging-in $\epsilon$, we have $ N_{m}\leq \frac{16R^2}{\epsilon^2} (h_{m,1} - \epsilon)$.
Following Eq. \eqref{eq:upper_bound_h_1_cg_lemma} we have 
\begin{align}
    &\mathbb{E}[N_{m}] \nonumber \\ &\leq  \frac{16R^2}{\epsilon^2} (\mathbb{E}[h_{m,1}] - \epsilon) \nonumber \\
    &\leq  \frac{16R^2}{\epsilon^2} \eta_m  \sqrt{\epsilon} ~ \left( \sqrt{K} \left(\frac{n M_{mK}}{\delta}\right) + K L_{mK} \right) \nonumber \\  &+ \frac{16R^2}{\epsilon^2} \eta_m^2 \left( K \left(\frac{n M_{mK}}{\delta}\right)^2 + K^2 L_{mK}^2 \right) + \frac{16R^2}{\epsilon^2}\eta_m D \sqrt{1-\frac{G_{m-1}}{G_{m}}} \left( \sqrt{K} \left(\frac{n M_{mK}}{\delta}\right) + K L_{mK} \right).
\end{align}
    Thus, overall on all blocks, we obtain 
    \begin{align}
       & \mathbb{E} \left[ \sum_{m=1}^{\frac{T}{K}} N_m \right] \nonumber \\
   & \leq  \frac{16R^2}{\epsilon^2}\Biggl (\sum_{m=1}^{\frac{T}{K}}  \eta_m  \sqrt{\epsilon} ~ \left( \sqrt{K} \left(\frac{n M_{mK}}{\delta}\right) + K L_{mK} \right)  + \eta_m^2 \left( K \left(\frac{n M_{mK}}{\delta}\right)^2 + K^2 L_{mK}^2 \right) \nonumber \\
    & + \sqrt{\sum_{m=1}^{\frac{T}{K}}\eta_m^2 \left( K \left(\frac{n M_{mK}}{\delta}\right)^2 + K^2 L_{mK}^2 \right)\Biggl)}D \sqrt{\log \frac{G_{T/K}}{G_1}}\Biggl) \nonumber \\
    & \leq \frac{16R^4}{\epsilon^2} \left ( \frac{\sqrt{\epsilon}}{R} T^{1/4} + 1 + \sqrt{\log \frac{G_{T/K}}{G_1}}\right)
    \end{align}

Where the last inequality follows because we take $\eta_m = \frac{2R}{K(\frac{nM_{mK}}{\delta} + L_{mK}\sqrt{K})}.$
\subsection{Auxiliary Lemmas}
\iffalse
\begin{lemma}
\label{lemma:ftrl-inequality}
Let $w_1,w_2,...$ be the sequence of vectors produced by FTRL. Then, for all $u\in S$ we have $$\sum_{t+1}^{T}(f_t(w_t)-f_t(u))\leq \sum_{1}^{T}(f_t(w_t)-f_t(w_{t+1}))+R_T(u)$$
where $\{R_t\}_{1\leq t \leq T}$ is a non-decreasing functional sequence.
\end{lemma}
\begin{proof}
\begin{align}
    \tilde{f_t}=f_t+R_t-R_{t-1} \,\,for \,T\geq 1
\end{align}
Observe that running $f_1,....,f_T$ on FoReL is equivalent for running $\tilde{f_1},...\tilde{f_T}$ on FTL.
\begin{align}
    &\sum_{1}^{T}(\tilde{f}_t(w_t)-\tilde{f}_t(u))\leq \sum_{1}^{T}(\tilde{f}_t(w_t)-\tilde{f}_t(w_{t+1}))\\
    \implies  &\sum_{1}^{T} (f_t(w_t)-f_t(u)) \leq \sum_{1}^{T}(f_t(w_t)-f_t(w_{t+1}))\\
    &+\sum_{1}^{T} \underbrace{[(R_t(u)-R_{t-1}(u)]}_{\geq 0}-\sum_{1}^{T}\underbrace{[R_t(w_{t+1})-R_{t-1}(w_{t+1})]}_{\geq 0}\\
    &\leq  \sum_{1}^{T}(f_t(w_t)-f_t(w_{t+1}))+\sum_{1}^{T} [(R_t(u)-R_{t-1}(u)]\\
    &=\sum_{1}^{T}(f_t(w_t)-f_t(w_{t+1}))+R_T(u)
\end{align}
\end{proof}
\fi

\begin{lemma}
\label{ftrl-drift}
    Under Algorithm \ref{alg:VRBCO}, the following holds
	\begin{eqnarray*}
		\Vert x_{m}^*-x_{m+1}^* \Vert \leq \eta_m  \Vert \nabla_{m} \Vert + D\sqrt{(1-\frac{G_{m-1}}{G_m})}
	\end{eqnarray*}
\end{lemma}
\begin{proof}
    \begin{eqnarray}
       &&\Vert x_{m}^*-x_{m+1}^* \Vert ^2/ \eta_{m} \leq F_{m}(x_{m}^*) - F_{m}(x_{m+1}^*) 
      \leq  \underbrace{F_{m-1}(x_{m}^*)- F_{m-1}(x_{m+1}^*)}_{\leq 0} +  \nabla_{m}^{\top} (x_{m}^* - x_{m+1}^*) \nonumber \\ 
      && + \Vert x_{m}^*-x_{1}^* \Vert ^2(\frac{1}{\eta_{m}}-\frac{1}{\eta_{m-1}})  \nonumber \\
      &\implies&\Vert x_{m}^*-x_{m+1}^* \Vert ^2\leq  \eta_m  \Vert \nabla_{m} \Vert  \Vert (x_{m}^* - x_{m+1}^*) \Vert + D^2(1-\frac{G_{m-1}}{G_{m}})\nonumber \\
      &\implies&\Vert x_{m}^*-x_{m+1}^* \Vert \leq \eta_m  \Vert \nabla_{m} \Vert + D\sqrt{(1-\frac{G_{m-1}}{G_m}}). \label{eq:absolut_dist}
    \end{eqnarray}
\end{proof}
\begin{lemma} \label{lemma:expectation_gradient}
    %Define $\mathbf{g}_t = \frac{n}{\delta} f_t(\mathbf{y}_t) \mathbf{u}_t$ which is unbiased estimator of $\nabla {\hat{f}}_{t,\delta} (\mathbf{x}_t)$, $s=(m-1)K$ and $\hat{\mathbf{g}}_m = \sum_{t=s+1}^{s+K} \mathbf{g}_t$ Then, for all $K, \delta, n$ the following holds
    For any iteration (block) $m$ of the outer-loop in Algorithm \ref{alg:VRBCO} it holds that
    \begin{align}
        \mathbb{E} \left[  \Vert \nabla_m\Vert  \right]^2 \leq \mathbb{E} \left[  \Vert \nabla_m \Vert ^2 \right] \leq K \left(\frac{n M_{mK}}{\delta}\right)^2 + K^2 L_{mK}^2. \nonumber
    \end{align}
\end{lemma}
\begin{proof}
    \begin{align}
        \mathbb{E} \left[  \Vert \nabla_m \Vert ^2 \right] & =   \mathbb{E} \left[ \Vert \sum\limits_{s=1}^{K} \nabla_{m,s} \Vert ^2 \right] \nonumber \\
        = & \mathbb{E} \left[\sum\limits_{s=1}^{K}  \Vert \nabla_{m,s} \Vert ^2 + \sum\limits_{s_2=1}^{K}\sum\limits_{s_1=1, s_1 \neq s_2}^{K}\nabla_{m,s_1}^{\top} \nabla_{m,s_2} \right] \nonumber \\
        = & \mathbb{E} \left[\sum\limits_{s=1}^{K}  \Vert  \nabla_{m,s} \Vert ^2 \right] + \sum\limits_{s_2=1}^{K}\sum\limits_{s_1=1, s_1 \neq s_2}^{K}\E[\nabla_{m,s_1}^{\top} \nabla_{m,s_2}]. \nonumber 
    \end{align}	
    Since, conditioned on the iterate $x_{m-1}$, $\forall s_1 \neq s_2$  $\nabla_{m,s_1}$, $\nabla_{m,s_2}$ are independent random vectors, we have
    {\begin{align*}
        &\mathbb{E} \left[  \Vert \nabla_m \Vert ^2 \right] = \mathbb{E} \left[\sum\limits_{s=1}^{K}  \Vert \nabla_{m,s} \Vert ^2 \right] + \sum\limits_{s_2=1}^{K}\sum\limits_{s_1=1, s_1 \neq s_2}^{K}\E[\nabla_{m,s_1}^{\top}] \E[\nabla_{m,s_2}] .  \nonumber
    \end{align*}}
    We have that for all $(m-1)K+1\leq t\leq mK, \Vert{\mathbb{E} [\nabla_t|x_{m}]}\Vert = \Vert{\nabla {\hat{f}}_{t,\delta} (\mathbf{x}_{m})}\Vert \leq L_{mK}$.
    Since $\max_{\mathbf{x} \in \mathcal{K}}  \Vert f(\mathbf{x}) \Vert  \leq M$, we also have $  \Vert \nabla_{m,s} \Vert  \leq \frac{n}{\delta}  \Vert f_t(\mathbf{y}_t) \Vert   \Vert \mathbf{u}_t \Vert  \leq \frac{nM_{mK}}{\delta}$, and thus,
    {\small\begin{align*}
        & \mathbb{E} \left[\sum\limits_{s=1}^{K}  \Vert \nabla_{m,s} \Vert ^2 \right]  
        + \sum\limits_{s_2=1}^{K}\sum\limits_{s_1=1, s_1 \neq s_2}^{K}\E[\nabla_{m,s_1}^{\top}] \E[\nabla_{m,s_2}] \\ 
        &\leq  K \left(\frac{n M_{mK}}{\delta}\right)^2 + \left(K^2 - K \right) L_{mK}^2 \leq K \left(\frac{n M_{mK}}{\delta}\right)^2 + K^2L_{mK}^2. \nonumber
    \end{align*}}
      Finally, the inequality $\mathbb{E} \left[  \Vert \nabla_m \Vert  \right]^2 \leq \mathbb{E} \left[  \Vert \nabla_m \Vert ^2 \right] $ stated in the lemma follows from using Jensen's inequality.
\end{proof}

\begin{lemma}
    \label{lemma:cg_epsilon_error_L_iterations} Given a function $F(\mathbf{x})$, $2$-smooth and $2$-strongly convex, and $\mathbf{x}_{1} \in \mathcal{X}_{\delta}$ such that $h_1 = F(x_{1}) - F(x^*) \leq \tilde{\epsilon}$, where $x^* = \argmin\limits_{x \in \mathcal{X}_{\delta}}F(x) $, Algorithm \ref{alg:Cg} produces a point $x_{L+1}\in\mathcal{X}_{\delta}$ such that $F(x_{L+1}) - F(x^*) \leq \epsilon$ after at most $L = \max \bigg{\{} \frac{16R^2}{\epsilon^2} (h_1 - \epsilon) , 
    ~ \frac{2}{\epsilon} (h_1 - \epsilon) \bigg{\}}$ iterations.
\end{lemma}
\begin{proof}
    For any iteration $\tau$ of Algorithm \ref{alg:Cg}, define $h_{\tau} =  F(x_{\tau}) - F(x^*)$ and denote $\nabla_{\tau} = \nabla F(x_{\tau})$. From the definition of $v_{\tau}$ (in Line 6 of Algorithm \ref{alg:Cg})and the convexity of $F(\cdot)$, it follows that
%     Due to $\forall \mathbf{x} \in \mathcal{K}_{\delta}$ exists $\nabla_{\tau}^{\top} (\mathbf{v}_{\tau}) \leq \nabla_{\tau}^{\top} (\mathbf{x})$, specifically $\mathbf{x}^*$, and from the convexity of $F(\mathbf{x})$, we have
    \begin{align}
        \nabla_{\tau}^{\top} (x_{\tau} - v_{\tau}) \geq & \nabla_{\tau}^{\top} (x_{\tau} - x^*) \nonumber \\
        \geq & F(x_{\tau}) - F(x^*) = h_\tau \label{eq:aux1_cg_lemma}
    \end{align}
    Now, we establish the convergence rate of Algorithm \ref{alg:Cg}. It holds that
    \begin{align}
        h_{\tau+1} = & F(x_{\tau+1}) -  F(x^*) \nonumber \\
        = & F(x_{\tau} + \sigma_{\tau}(v_{\tau} - x_{\tau})) - F(x^*) . \nonumber 
    \end{align}
    For our analysis we define the step-size $\hat{\sigma}_{\tau} = \min \Big{\{} \frac{ \nabla_{\tau}^{\top} (x_{\tau} - v_{\tau})}{8R^2}, 1 \Big{\}}$. %Even though Algorithm \ref{alg:Cg} $\sigma$, which is the best, $\sigma_{\tau} = \argmin\limits_{\sigma \in [0,1]} \big{\{} F(\mathbf{x}_{\tau} + \sigma(\mathbf{v}_{\tau} - \mathbf{x}_{\tau})) \big{\}}$. Therefore,
    Since $\sigma_{\tau}$ is chosen via line-search, we have that
    \begin{align}
        h_{\tau+1} = &  F(x_{\tau} + \sigma_{\tau}(v_{\tau} - x_{\tau})) - F(x^*) \nonumber \\
        \leq & F(x_{\tau} + \hat{\sigma}_{\tau}(v_{\tau} - x_{\tau})) - F(x^*). \nonumber
    \end{align}
    The above step was done to ensure that we can use our own $\hat{\sigma}_{\tau}$ to reach our bound.
    
    Since $F(x)$ is $2$-smooth it holds that 
    \begin{align*}
    F(x_{\tau} + \hat{\sigma}_{\tau}(v_{\tau} - x_{\tau})) &\leq F(x_{\tau}) + \hat{\sigma}_{\tau} \nabla_{\tau}^{\top} (v_{\tau} - x_{\tau}) \\
    &+ \hat{\sigma}_{\tau}^2  \Vert v_{\tau} - x_{\tau} \Vert ^2, 
    \end{align*}
    and we obtain
    \begin{align}
        h_{\tau+1} \leq & h_{\tau} + \hat{\sigma}_{\tau}^2 (2R)^2 - \hat{\sigma}_{\tau} \nabla_{\tau}^{\top} (x_{\tau} - v_{\tau}).   \nonumber
    \end{align}
    We now consider several cases.
    \\ \emph{Case 1}: If $\nabla_{\tau}^{\top} (x_{\tau} - v_{\tau}) \leq \epsilon$ for some $\tau < L$, the algorithm will stop after less than $L$ iterations. Moreover, from Eq. \eqref{eq:aux1_cg_lemma} we have $h_{\tau} \leq \epsilon$.
    \\ \emph{Case 2}: Else, $\nabla_{\tau}^{\top} (x_{\tau} - v_{\tau}) \geq \epsilon$ for all $\tau < L$. We have 2 cases: \\
    \emph{Case 2.1}: If $\nabla_{\tau}^{\top} (x_{\tau} - v_{\tau}) \geq 8R^2$ then $\hat{\sigma}_{\tau} = 1$ and we have
    \begin{align}
        h_{\tau+1} \leq & h_{\tau} + \hat{\sigma}_{\tau}^2 (2R)^2 - \hat{\sigma}_{\tau} \nabla_{\tau}^{\top} (x_{\tau} - v_{\tau})   \nonumber \\
        \leq & h_{\tau} - \frac{ \nabla_{\tau}^{\top} (x_{\tau} - v_{\tau})}{2} .\nonumber
    \end{align}
        \\ \emph{Case 2.2}: Else, $ \nabla_{\tau}^{\top} (x_{\tau} - v_{\tau}) \leq 8R^2$, and then $\hat{\sigma}_{\tau} = \frac{ \nabla_{\tau}^{\top} (x_{\tau} - v_{\tau})}{8R^2}$, and we have
    \begin{align}
        h_{\tau+1} \leq & h_{\tau} + \hat{\sigma}_{\tau}^2 (2R)^2 - \hat{\sigma}_{\tau}  \nabla_{\tau}^{\top} (x_{\tau} - v_{\tau})   \nonumber \\
        \leq & h_{\tau} - \left( \frac{\nabla_{\tau}^{\top} (x_{\tau} - v_{\tau})}{4R} \right)^2. \nonumber
    \end{align}
    From both cases, we have
    \begin{align}
        h_{\tau+1} & \leq h_{\tau} - \min \bigg{\{} \left( \frac{ \nabla_{\tau}^{\top} (x_{\tau} - v_{\tau})}{4R} \right)^2, \frac{ \nabla_{\tau}^{\top} (x_{\tau} - v_{\tau})}{2} \bigg{\}} \nonumber \\
        \leq & h_{1} - \tau \min_{i = 1, \dots, \tau} \bigg{\{} \left( \frac{ \nabla_{\tau}^{\top} (x_{i} - v_{i})}{4R} \right)^2, \frac{ \nabla_{\tau}^{\top} (x_{i} - v_{i})}{2} \bigg{\}} \nonumber \\
        & \leq  h_{1} - \tau \min \bigg{\{} \left( \frac{ \epsilon}{4R} \right)^2, \frac{ \epsilon}{2} \bigg{\}}.
    \end{align}
    Thus, for all cases, after a maximum of $L$ iterations, when
    \begin{align}
        L = \max \bigg{\{} \frac{16R^2}{\epsilon^2} (h_1 - \epsilon) , 
        ~ \frac{2}{\epsilon} (h_1 - \epsilon) \bigg{\}}, \nonumber
    \end{align}
    we obtain $ h_{L+1} \leq \epsilon$.

    So, when the loop naturally terminates, the condition is met - even if the loop doesn't terminate, we need to run it for a maximum of $L$ times to ensure that the condition is met. 
\end{proof}
\section{Proof of Theorem \ref{ftpl_perf_thm}}
\label{sec:ftpl_pf}
The following pseudocode, given in Algorithm \ref{alg:ftpl_adaptive}, describes the proposed FTPL-based projection-free algorithm for the OLO problem with constraints. 
\begin{algorithm}[ht]
\caption{Adaptive FTPL with Adversarial Constraints}
\label{alg:ftpl_adaptive}
\begin{algorithmic}[1]
\REQUIRE Compact decision set $\mathcal{X} \subseteq \mathbb{R}^N,$ parameter $V= \sqrt{T}$, linear cost and constraint functions with coefficients $\{f_t, g_t\}_{t \geq 1}$, Gaussian Perturbation $\mathcal{N}(0, I_N)$
\STATE \textbf{Initialize:} Virtual queue $Q(0) = 0$, Cumulative surrogate loss vector $L_0 = \mathbf{0}$
\FOR{$t = 1$ : $T$}
    \STATE Calculate adaptive scaling parameter: 
    \begin{equation*}
        \eta_t = \sqrt{\frac{1 + \sum_{\tau=1}^{t-1} \|\hat{f}_\tau\|_2^2}{N}}
    \end{equation*}
    \STATE Sample perturbation vector $p_t \sim \mathcal{N}(0, I)$
    \STATE The learner plays $x_t = \arg\min_{x \in \mathcal{X}} \langle L_{t-1} + \eta_t p_t, x \rangle$ 
    \STATE The adversary reveals cost vector $f_t$ and constraint vector $g_t$
    \STATE Update the virtual queue: $Q(t) = \left(Q(t-1) + \langle g_t, x_t \rangle \right)^+$
    \STATE Construct surrogate loss vector: $\hat{f}_t = V f_t + 2 Q(t-1) g_t$
    \STATE Update cumulative surrogate loss: $L_t = L_{t-1} + \hat{f}_t$
\ENDFOR
\end{algorithmic}
\end{algorithm}

\iffalse
\paragraph{Connection between GBPA and FTPL.} 
The Gradient-Based Prediction Algorithm (GBPA) provides a general framework for online learning by defining actions as the gradient of a potential function. The classical Follow The Perturbed Leader (FTPL) algorithm can be viewed precisely as an instance of GBPA where the potential function $\tilde{\Phi}_t$ is defined via stochastic smoothing. Specifically, if we define $\tilde{\Phi}_t(G) = \mathbb{E}_{z \sim \mathcal{D}}[ \max_{x \in \mathcal{X}} \langle x, G + \eta_t z \rangle ]$, the gradient $\nabla \tilde{\Phi}_t(G)$ yields the expected action of FTPL. Because the expected regret of GBPA on a stochastically smoothed potential is exactly equal to the expected regret of FTPL, we can use these algorithms interchangeably for linear optimization tasks. Notably, FTPL is a \emph{projection-free} policy which requires access to only a Linear Optimization Oracle (LOO) over the decision set $\mathcal{X}$.
\fi

\paragraph{Regret and Constraint Violation Analysis:}
Recall that we consider the online linear optimization setting with linear cost and constraint functions $f_t(x) = \langle f_t, x \rangle$ and $g_t(x) = \langle g_t, x \rangle$ such that $\|f_t\|, \|g_t\| \leq 1, \ \forall t \geq 1,$ and the action $x$ belongs to some compact set $\mathcal{X}$. Our analysis closely follows the derivation in Section \ref{gen_decomp}.

% Recall the standard FTPL regret bound achieved with Gaussian perturbations for linear cost functions \citep`[Theorem 1.10]{abernethy2016perturbation}:
% \begin{equation}
% \mathbb{E}\text{Regret}_T \leq 2 \sqrt{1 + \sum_{t=1}^{T} \|g_t\|^2}.
% \end{equation}

Define the virtual queue for Cumulative Constraint Violation (CCV) as:
\begin{equation} \label{q-ev-eqn}
Q(t) = \left(Q(t-1) + \langle g_t, x_t \rangle \right)^+
\end{equation}
and choose the quadratic Lyapunov function $\Phi(x) = \frac{x^2}{V}$, where $V > 0$ is a constant to be determined later. Eqn.\ \eqref{q-ev-eqn} yields the following drift inequality:
\begin{equation}
Q^2(t) \leq Q^2(t-1) + 1 + 2 Q(t-1) \langle g_t, x_t \rangle.
\end{equation}
Let $x^*$ be a feasible solution to the constraints. Adding $V(f_t(x_t)-f_t(x^\star))$ to both sides of the above and then summing up the resulting inequalities, we obtain the following regret decomposition inequality:
\begin{eqnarray}
    Q^2(t) + V \sum_{\tau=1}^t \big(f_\tau(x_\tau)-f_\tau(x^\star)\big) \leq  t+ \sum_{\tau=1}^t \big(\hat{f}_\tau(x_\tau)-\hat{f}_\tau(x^\star)\big),
\end{eqnarray}
where we have defined the surrogate linear cost function for round $t$ as:
\begin{equation}
\hat{f}_t(x) = \langle V f_t + 2 Q(t-1) g_t, x \rangle.
\end{equation}

Finally, running the FTPL policy with Gaussian noise and using the standard adaptive regret bound, given by \citep[Theorem 1.10]{abernethy2016perturbation}, on these surrogate costs, we obtain the following regret bound on the surrogate cost functions:
\begin{equation}
\text{Regret}'_t \leq 2 \sqrt{1 + \sum_{\tau=1}^{t} \left(V^2 + 4 Q^2(\tau-1)\right)} 
\leq 2 + V \sqrt{t} + 4 \sqrt{\sum_{\tau=1}^{t-1} Q^2(\tau)}.
\end{equation}
By scaling the Gaussian perturbation $p_t$ at each round $t$ by the adaptive parameter $\eta_t = \sqrt{(1 + \sum_{\tau=1}^{t-1} \|\hat{f}_\tau\|_2^2) / N}$, the algorithm dynamically adjusts to the magnitude of the observed surrogate losses.

Combining this with the Lyapunov drift yields the following fundamental regret-decomposition inequality:
\begin{equation}
\label{eq:original_ineq}
\mathbb{E}Q^2(t) + V \, \mathbb{E}\text{Regret}_t 
\leq 2 + t + V \sqrt{t} + 4 \sqrt{\sum_{\tau=1}^{t-1} \mathbb{E}Q^2(\tau)}.
\end{equation}

Following an inductive analysis similar to \cite{sinha2024optimal}, this decomposition guarantees sublinear bounds, which we rigorously show in the following.

Summing from $t=1$ to $T$ we have,
\begin{align} \label{Q-decomp-2}
    \sum_{t=1}^T\mathbb{E}Q^2(t) + V\sum_{t=1}^T\mathbb{E} \textrm{Regret}_t \leq VT^{3/2} + 2T + T^2 + 4 T\sqrt{\sum_{t=1}^T \mathbb{E}Q^2(t)}.
\end{align}

Let us now define the variable $R(T) := \sqrt{\sum_{t=1}^T \mathbb{E}Q^2(t)}.$ Note that we trivially have $\mathbb{E} \textrm{Regret}_t \geq -T$.  Plugging in this bound in inequality \eqref{Q-decomp-2}, we conclude:
\begin{align*}
    R^2(T) \leq VT^2 + T + T^2 + 4VT^{3/2} + 4 TR(T).
\end{align*}
Noticing that the above inequality is of the form $x^2 \leq ax + b$ where $x\equiv R(T),$ and because this implies $x\leq a + \sqrt{b}$, we obtain the following upper bound on $R(T)$:
\begin{align}
\label{eq:bound-R(t)}
    R(T) \leq \sqrt{V}T + \sqrt{T} + 2\sqrt{V}T^{3/4} + 5 T.
\end{align}
\iffalse
\begin{remark}
\label{rem:neg-reg}
Note that in the unconstrained problem $\mathbb{E} \textrm{Regret}_t (\pi^\star) \geq 0$ as the comparator policy $\pi^\star$ is, by definition, the \textit{best policy} in expectation. However, in the constrained setting, because we are now limited to choosing a benchmark policy from the set of all feasible policies $\Pi^\star$, the non-negativity of regret may no longer hold. This hints at a potential trade-off between feasibility and regret.
\end{remark}
\fi

We further note that inequality \eqref{eq:original_ineq} can be rewritten in term of the variables $R(T)$ as follows:
\begin{align} \label{q-reg-ineq0}
\mathbb{E}Q^2(T) + V\mathbb{E} \mathsf{Regret}_T \leq V\sqrt{T} + 2 + T + 4R(T).
\end{align}
Plugging in the upper bound on $R(T)$ from \eqref{eq:bound-R(t)} into the above inequality, we obtain
\begin{align} \label{Q-reg-ineq1}
\mathbb{E}Q^2(T) + V\mathbb{E} \textrm{Regret}_T \leq 4V\sqrt{T} + 2 + T +  4\sqrt{V}T + 8\sqrt{T} + 8\sqrt{V} T^{3/4} + 20T.
\end{align}
Hence, using $\mathbb{E}(Q^2(T))\geq 0,$ we obtain the following regret bound:
\begin{align*}
    \mathbb{E} \mathsf{Regret}_T \leq O\bigg(\max(\sqrt{T}, \sqrt{\frac{T}{V}}, \frac{T^{3/4}}{\sqrt{V}}, \frac{T}{V})\bigg).
\end{align*}
Furthermore, substituting the trivial regret lower bound $\mathbb{E} \textrm{Regret}_T \geq -T$ into \eqref{Q-reg-ineq1} and using Jensen's inequality $\mathbb{E}Q^2(T) \geq (\mathbb{E}(Q(T)))^2,$ we obtain the following CCV bound
%\cmt{elaborate the following \CCV~bound.}
\begin{eqnarray} \label{q-bd-no-assump}
    \mathbb{E}Q(T)\leq O\bigg(\max(\sqrt{VT},\sqrt{V}(T)^{1/4},  (V)^{1/4}\sqrt{T} , V^{1/4} T^{3/8}, \sqrt{T})\bigg).
\end{eqnarray}
Choosing $V = \sqrt{T},$ we conclude $ \mathbb{E} \mathsf{Regret}_T = O(T^{3/4}),~ \mathbb{E}\mathsf{CCV}_T = O(T^{3/4}).$
\section{Budget Constrained Benchmark}
\label{sec:budget_constraints}
Let $B_T$ denote the long-term budget and let us denote the set of all budget-feasible benchmarks by $\mathcal{X}^\star$, i.e.,

$$
\mathcal{X}^\star = \big\{x \in \mathcal{X}: \sum_{t=1}^T g_t^+(x) \leq B_T\big\}.
$$

As before,
\[
\Phi(Q(t)) - \Phi(Q(t-1)) \leq \Phi'(Q(t)) g_t(x_t)^+
\]
\[
\Phi(Q(t)) - \Phi(Q(t-1)) + Vf_t(x_t) -  V f_t(x^\star) - \Phi'(Q(t)) g_t^+(x^\star) \leq Vf_t(x_t) + \Phi'(Q(t)) g_t^+(x_t) -  V f_t(x^\star) - \Phi'(Q(t)) g_t^+(x^\star)
\]
where we added $Vf_t(x_t) -  V f_t(x^\star) - \Phi'(Q(t)) g_t^+(x^\star)$ and re-arranged the terms on the RHS to make the structure of the surrogate loss apparent. 

Noting that, $\hat{f}_t(x) = Vf_t(x)+ \Phi'(Q(t))g_t^+(x).$
\[
\Phi(Q(t)) - \Phi(Q(t-1)) + Vf_t(x_t) -  V f_t(x^\star) \leq \hat{f}_t(x_t) -  \hat{f}_t(x^\star) + \Phi'(Q(t)) g_t^+(x^\star)
\]

Further summing from $t=1$ to $T$,
\[
\Phi(Q(T)) - \Phi(Q(0)) + V\textrm{Regret}_T(x^\star) \leq \textrm{Regret}'(x^\star) + \sum_{t=1}^T \Phi'(Q(t)) g_t^+(x^\star)
\]
\[
\Phi(Q(T)) - \Phi(Q(0)) + V\textrm{Regret}_T(x^\star) \leq \textrm{Regret}'(x^\star) + \Phi'(Q(T))\sum_{t=1}^T  g_t^+(x^\star)
\]

\[
\Phi(Q(T)) - \Phi(Q(0)) + V\textrm{Regret}_T(x^\star) \leq \textrm{Regret}'(x^\star) + \Phi'(Q(T))B_T.
\]

Noting that,
\[ \textrm{Regret}_T' \leq  T^{3/4} (V+\Phi'(Q(T))) \log T.\]

Then we get,

\[
\Phi(Q(T)) - \Phi(Q(0)) + V\textrm{Regret}_T(x^\star) \leq VT^{3/4}\log T + \Phi'(Q(T))(T^{3/4}\log T+B_T)
\]
This is the new regret-decomposition inequality, which extends Eq. \eqref{reg-decomp} to the arbitrary budget setting. 
Calculations similar to those in our main paper with the power-law potential would yield,
\[
\textrm{Regret}_T(x^\star) = \Tilde{O}(T^{3/4})
\]
\[
Q(T) = \Tilde{O}(T^{3/4}) + B_T\log T
\]

It should be noted that for $B_T \leq T^{3/4}$, the strengthening of the benchmark does not make the violation worse. 

\section{Experiments: Online Shortest Path Problem with Constraints} \label{expts}
%\cmt{Dhruv and Aprameyo, please finish this.}
%\section{Experiments} \label{expts}
%In this section, we detail the experiments we conducted to prove the practical utility of our methodology. First, we begin by introducing the problem setting.
\paragraph{Setup:} We consider a constrained version of the online shortest path problem \citep{hazan2022introduction}, where the length corresponds to the latency and the constraint is on the long-term cumulative bandwidth across the path. In this problem, on each round, the online algorithm first selects a route connecting a source $s$ to a destination $d$ on a graph $G(V,E)$. The latency and bandwidth of each edge vary across rounds, reflecting dynamic network conditions. The objective is to minimize the cumulative latency subject to a long-term lower bound on the cumulative bandwidth. 
% We take the problem of navigation through the roads in a city while trying to minimise the time delay and fuel consumption. The abstract version of this problem is the Online Shortest Paths on Graph problem as the locations can be modelled as vertices and the roads can be modelled as edges.
Formally, 
%we are given a directed graph $G = (V, E)$ and a source-destination pair $s, d \in V.$ Then at 
the following sequence of events takes place on the $t$\textsuperscript{th} round: 
\begin{enumerate}
    \item The algorithm first chooses a route (randomly or otherwise) $p_t \in P_{s,d}$, where $P_{s,d}$ is the set of all $s-d$ routes in the graph. 
    \item  A latency of $\tau_e(t)$ and a bandwidth of $l_e(t)$ is chosen by an adversary for each edge $e \in E$. 
    \item The algorithm incurs a latency cost of $\sum_{e \in p_t} \tau_e(t)$ and a bandwidth cost of $-\sum_{e \in p_t} l_e(t)$ on round $t.$
\end{enumerate}
%Our specific problem setting is similar except that we not only have costs $f_t$ but also constraints $g_t$. This corresponds to modelling the delay of a path as the cost and the estimated fuel consumption when travelling across it as the constraint.

%We associate each path $x$ with a vector $\{0,1\}^{|E|}$, where $x(i)$ represents the presence of the $i$th edge in the path. 
We represent each route $p$ by its corresponding $|E|$-dimensional binary incidence vector where $p_e=1$ if the edge $e \in E$ belongs to the route or $p_e=0$ otherwise. On each round, our online policy returns an element from the convex hull of $P_{s,d}$, also known as the unit flow polytope \citep{hazan2022introduction}. We use Dijkstra's algorithm for computing the weighted shortest route in   Eqn.\ \eqref{v-def}. It is well-known that any element in the unit flow polytope can be efficiently decomposed into a convex combination of at most $|E|$ number of $s-d$ routes using the flow decomposition lemma \citep[Lemma 2.20]{williamson2019network}. These convex combinations can be used to randomly select a single route on each round, incurring the same expected cost. The experiments are performed on a quad-core CPU with 8 GB RAM.

%Note that our algorithms do not return single paths; instead, they return a convex combination of paths. This can also be interpreted as a distribution over paths. Thus, our decision set is the convex hull of the set of all $s-t$ paths. Notice that each element of this decision set can also be considered a unit flow. Thus this decision set is essentially the flow polytope which is defined by $O(|E|)$ linear equations : those for ensuring positivity of flow and those for conservation of in-flow and out-flow for every vertex other than source/sink (which will have unit inflow/outflow respectively).

%It is also possible to decompose the unit flow to a distribution of at most $|E|$ paths in polynomial time using the flow decomposition algorithm.

\paragraph{Dataset:} For the experiments, we construct a synthetic dataset from the raw data collected from RIPE Atlas - a global network measurement platform \citep{staff2015ripe}. The dataset provides real-time internet measurements, including latency and bandwidth, with temporal variations introduced through random scaling. We construct a graph with $n=191$ nodes and $m=1200$ edges where each edge is characterized by its latency and bandwidth values. We consider the horizon length to be $T=1600$ for our experiments.
%The resulting latency and bandwidth matrices provide a realistic foundation for evaluating the performance of network algorithms under diverse and dynamic scenarios. 
%We defer the details about the dataset to Section \ref{data collection} in the Appendix.
\paragraph{Results:} We compare our algorithm with three other competing algorithms which consider similar problem settings. The first is the COCO algorithm proposed by \citet{sinha2024optimal}, and the second is the Algorithm 3 of \citet{garber2024projectionfree} and finally we have the algorithm proposed by \citet{wang2025revisitingprojectionfreeonlinelearning}. The parameters have been set as specified in the corresponding papers.
%\cite{garber2024projectionfree} also proposed another algorithm (Algorithm 4) but it cannot be directly compared with our algorithm as its regret and violation guarantees do not hold for each timestep and instead only holds when the entire sequence of cost and constraint functions is considered.
We show the CCV plot in Figure \ref{fig:Figure 1}, Regret plot in Figure \ref{fig:Figure 2}, and execution times in Figure \ref{fig:Figure 3} for all three algorithms.
\begin{figure}[ht]
    \centering
    \begin{minipage}[b]{0.48\textwidth}
        \centering
        \includegraphics[width=\linewidth]{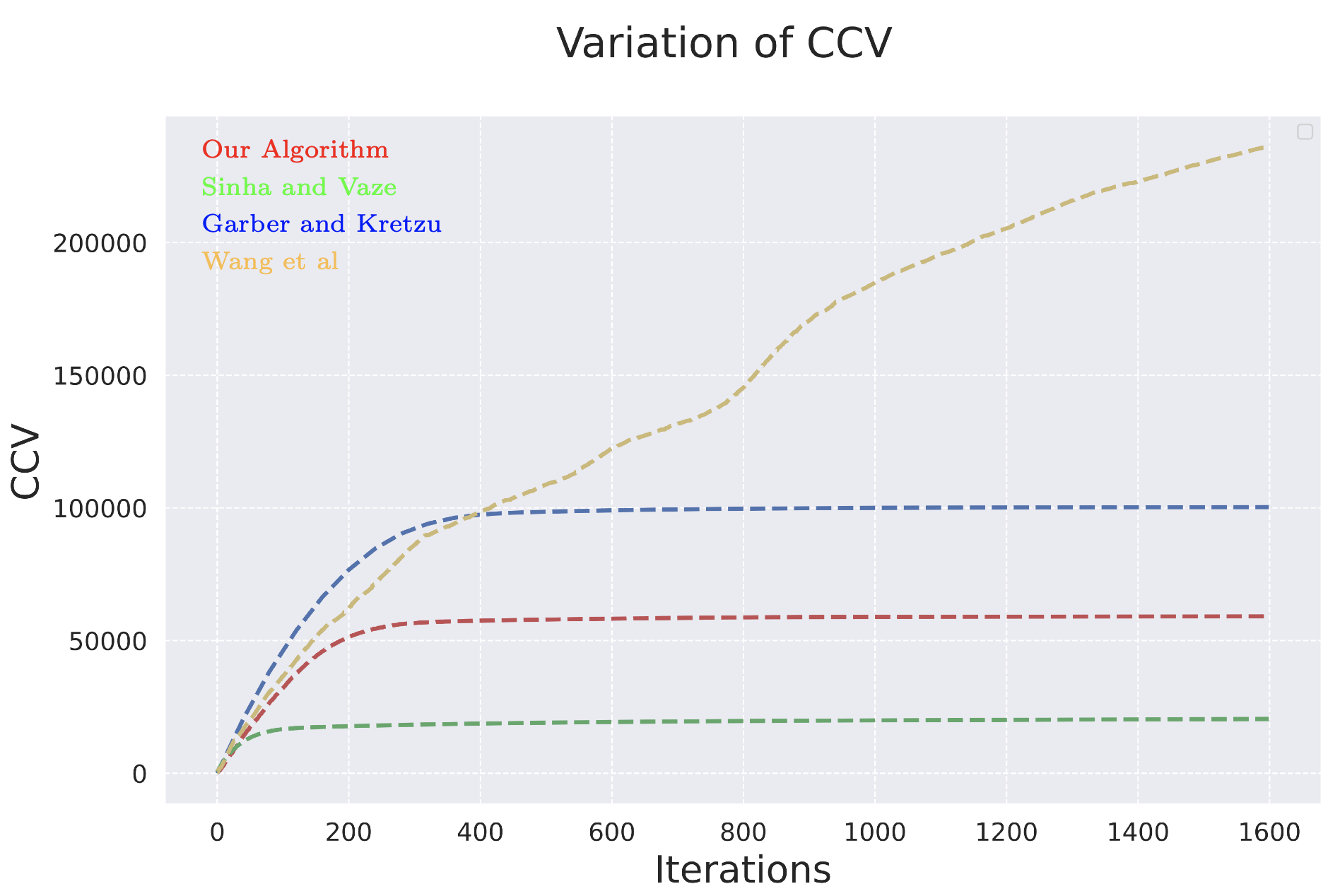}
        \caption{CCV comparison between our policy, Algorithm 1 of \citet{sinha2024optimal},  Algorithm 3 of \citet{garber2024projectionfree} and Algorithm 1 of \citet{wang2025revisitingprojectionfreeonlinelearning}.}
        \label{fig:Figure 1}
    \end{minipage}
    \hfill
    \begin{minipage}[b]{0.48\textwidth}
        \centering
        \includegraphics[width=\linewidth]{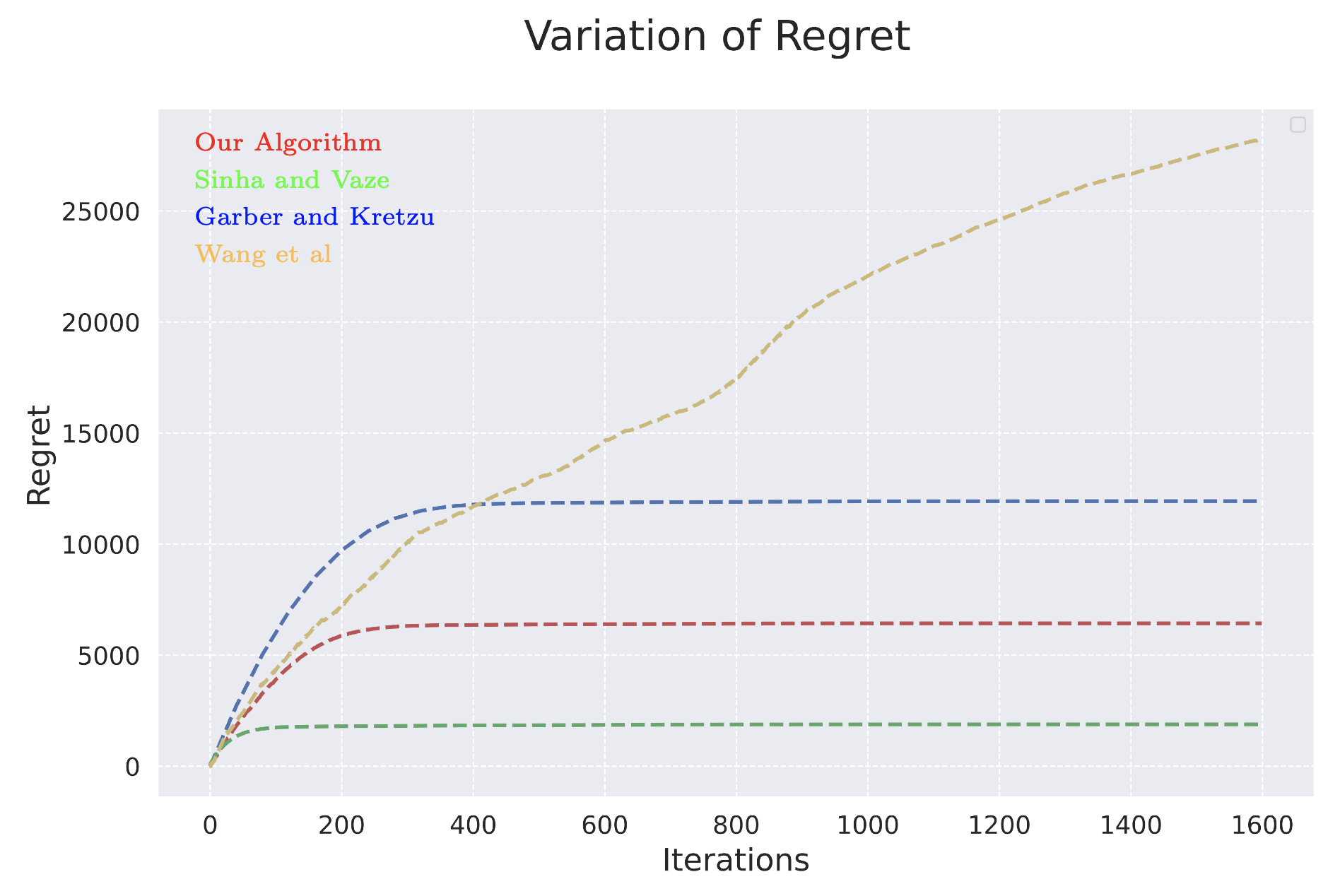}
        \caption{Regret comparison between our policy, Algorithm 1 of \citet{sinha2024optimal},  Algorithm 3 of \citet{garber2024projectionfree} and Algorithm 1 of \citet{wang2025revisitingprojectionfreeonlinelearning}.}
        \label{fig:Figure 2}
    \end{minipage}
    %\caption{Performance of our policy compared to baselines.}
    \label{fig:combined}
\end{figure}

\iffalse
\begin{figure}[h]
    \centering
    \includegraphics[width=.7\linewidth]{images/ccv3.png}
    \caption{Comparing CCV achieved by our policy, Algorithm 1 of \citet{sinha2024optimal},and Algorithm 3 of \citet{garber2024projectionfree}}
    \label{fig:Figure 1}
 \end{figure}
%\cmt{Does OCG mean our policy? In that case, please highlight it.}
\begin{figure}[h]
    \includegraphics[width=.5\linewidth]{images/reg3.png}
    \caption{Regret achieved by our policy,\\ Algorithm 1 of \citet{sinha2024optimal},\\ and Algorithm 3 of \citet{garber2024projectionfree}}
    \label{fig:Figure 2}
 \end{figure}
\fi
 \begin{figure}[ht]
 \centering
    \includegraphics[width=.5\linewidth]{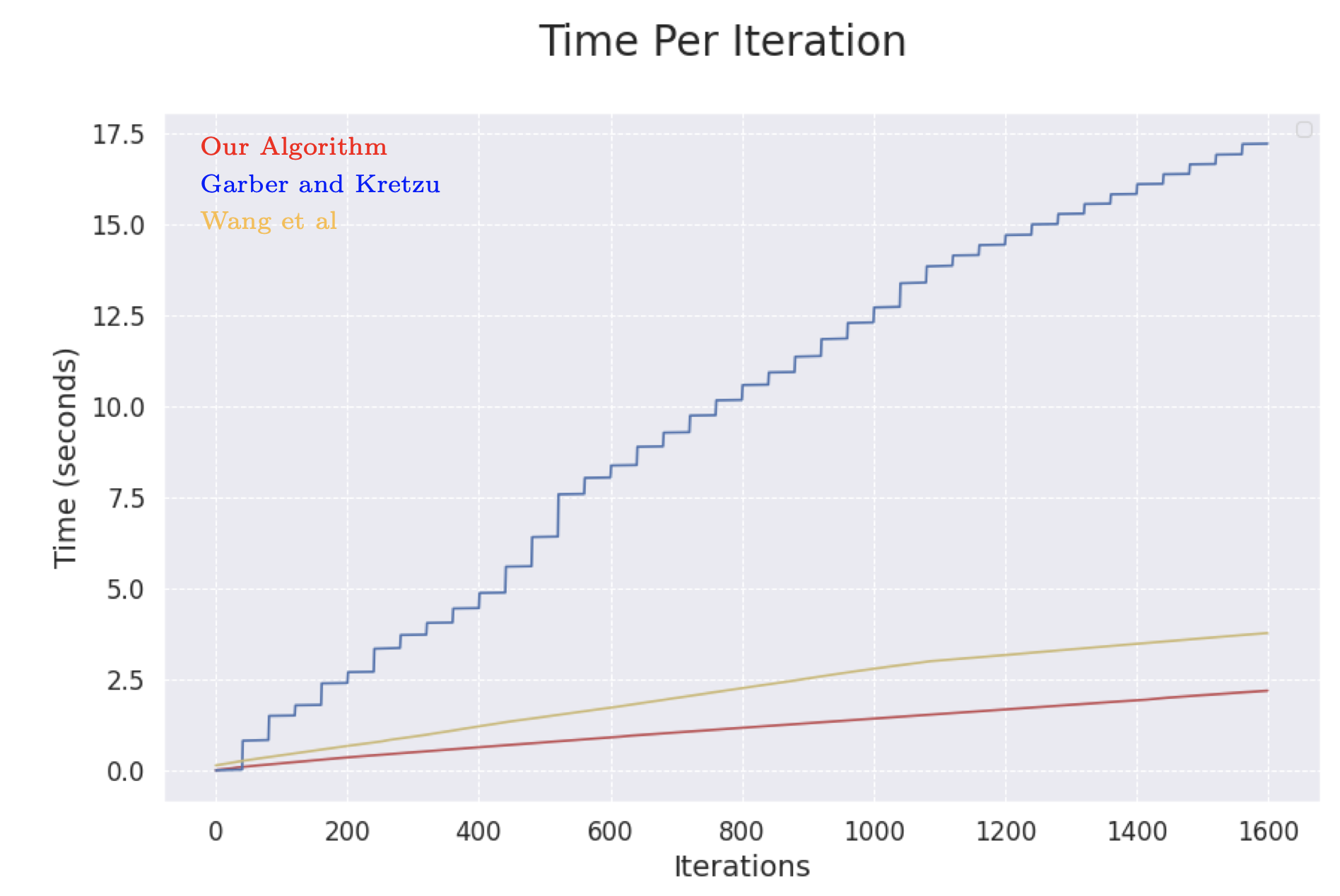}
    \caption{\small{Comparison of execution times of the proposed Algorithm \ref{our_alg} and Algorithm 3 of \citet{garber2024projectionfree} and also the algorithm proposed by \cite{wang2025revisitingprojectionfreeonlinelearning}. The run-time for \citet{sinha2024optimal}'s algorithm is two  orders of magnitude larger (more than $1000$s) and hence, is not shown on the figure. } }
    \label{fig:Figure 3}
 \end{figure}
 Figures \ref{fig:Figure 1} and \ref{fig:Figure 2} demonstrate that, in terms of regret and CCV, the algorithm proposed by \citet{sinha2024optimal} performs best when the computational cost is disregarded. This is not surprising because it is a computationally expensive projection-based algorithm whose theoretical guarantees of regret and CCV are much stronger ($O(\sqrt{T})$ and $\tilde{O}(\sqrt{T})$ respectively). However, this is more than compensated by the computational efficiency of our proposed Algorithm \ref{our_alg}. Furthermore, Algorithm \ref{our_alg} clearly outperforms the projection-free algorithm proposed in \citet{garber2024projectionfree} as demonstrated by the plots in Figure \ref{fig:Figure 2} and \ref{fig:Figure 3}. This is not surprising as the theoretical guarantees of \citet{garber2024projectionfree} are worse than ours. However, the performance of the algorithm proposed in \cite{wang2025revisitingprojectionfreeonlinelearning} may come as a surprise since it had comparable theoretical guarantees as ours and better than those in \cite{garber2024projectionfree}. As discussed previously, this is because the method in the paper relies on the impractical doubling trick. If one looks closely at the nature of the plot, it becomes clear that while the curves of the other algorithms reach a flat region, the curve corresponding to the algorithm in \cite{wang2025revisitingprojectionfreeonlinelearning} has a wavy nature and never really seems to reach a flat portion. This is due to the restarting technique that they employ where an the algorithm has to discard the information it has upto that point and start from scratch. Our observations match those in \cite{besson2018doubling} where they evaluated several versions of the doubling trick and concluded that performance is substantially worse than that of an intrinsically anytime algorithm, even one with less favorable theoretical guarantees. Moreover, this performance gap becomes larger each time the algorithm is restarted. 

 The projection-free algorithms mostly have similar computational efficiencies. The algorithm by \cite{garber2024projectionfree} has slightly worse computational efficiency since at one point it has to do convex optimization over a ball besides using LOO calls. However, compared to the algorithm by \cite{sinha2024optimal} which has to employ extremely computationally expensive steps, all three projection-free algorithms are much more efficient. 
%. Also, our algorithm has superior run-time performance than Algorithm 3 of \citet{garber2024projectionfree}, as evidenced by Figure \ref{fig:Figure 3}. 
%The run-time efficiency of our algorithm is explained by the fact that it only uses linear optimization, whereas Algorithm 3 of \citet{garber2024projectionfree} uses convex optimization. 

\FloatBarrier
%These encouraging results demonstrate the practicality of our algorithm. 

\subsection{Data collection}
\label{data collection}
Our data collection is done using public probes from RIPE Atlas, a global network measurement platform. Specifically, we collect HTTPS data that includes bandwidth and latency measurements between different network nodes. This data is gathered from a public probe, with the measurements taken between 20:30 and 21:30 on December 14th, 2024, and saved in the standard JSON format. The file contains fields such as IP addresses, timestamps, and corresponding latency and bandwidth values.

Once the data is collected, we proceed to construct a network graph. The graph is centered around a hub node, which connects to other nodes based on the latency and bandwidth data obtained from RIPE Atlas. Additional edges are then randomly added between other nodes, with the latency of each edge being the sum of the individual latencies to the central hub. The bandwidth of an edge is determined by the minimum bandwidth of the two nodes involved in the connection. This creates a network where each edge represents a connection between two nodes, characterized by their latency and bandwidth values.

To simulate real-world fluctuations and enhance the dataset, we introduce random scaling factors. Specifically, the latency values are scaled randomly between 0.5x and 1.5x, while the bandwidth values are scaled between 0.8x and 1.2x for each iteration. This variability mimics the dynamic nature of communication networks, ensuring that the dataset reflects more realistic network conditions. The scaled latency and bandwidth values are then used to construct latency and bandwidth matrices, which serve as the basis for the performance evaluation of the algorithms.
\end{document}